\documentclass[10pt,twocolumn,letterpaper]{article}

\usepackage{iccv}              

\usepackage[accsupp]{axessibility}

\usepackage{iccv}
\usepackage{times}
\usepackage{epsfig}
\usepackage{graphicx}
\usepackage{amsmath}
\usepackage{amssymb}
\usepackage{multirow}
\usepackage{makecell}
\usepackage{amssymb}

\definecolor{iccvblue}{rgb}{0.21,0.49,0.74}
\usepackage[pagebackref,breaklinks,colorlinks,allcolors=iccvblue]{hyperref}

\def\paperID{11675} 
\def\confName{ICCV}
\def\confYear{2025}

\title{Normal and Abnormal Pathology Knowledge-Augmented Vision-Language Model for Anomaly Detection in Pathology Images}

\author{
Jinsol Song\textsuperscript{1} \quad Jiamu Wang\textsuperscript{1} \quad Anh Tien Nguyen\textsuperscript{1} \quad Keunho Byeon\textsuperscript{1}\\
Sangjeong Ahn\textsuperscript{1} \quad Sung Hak Lee\textsuperscript{2} \quad Jin Tae Kwak\textsuperscript{1}\\
\textsuperscript{1}Korea University \quad \textsuperscript{2}The Catholic University of Korea\\
{\tt\small Jinsol Song: truetg@korea.ac.kr} \quad {\tt\small Jin Tae Kwak: jkwak@korea.ac.kr}
}

\begin{document}
\maketitle
\begin{abstract}
Anomaly detection in computational pathology aims to identify rare and scarce anomalies where disease-related data are often limited or missing. Existing anomaly detection methods, primarily designed for industrial settings, face limitations in pathology due to computational constraints, diverse tissue structures, and lack of interpretability. To address these challenges, we propose Ano-NAViLa, a \textbf{N}ormal and \textbf{A}bnormal pathology knowledge-augmented \textbf{Vi}sion-\textbf{La}nguage model for \textbf{Ano}maly detection in pathology images. Ano-NAViLa is built on a pre-trained vision-language model with a lightweight trainable MLP. By incorporating both normal and abnormal pathology knowledge, Ano-NAViLa enhances accuracy and robustness to variability in pathology images and provides interpretability through image-text associations. Evaluated on two lymph node datasets from different organs, Ano-NAViLa achieves the state-of-the-art performance in anomaly detection and localization, outperforming competing models.
\end{abstract}    
\section{Introduction}
Anomaly detection (AD) has been a long-standing research area dedicated to addressing the practical challenge of detecting anomalies, which are inherently scarce and difficult to collect compared to normal data. While AD can be considered as a binary classification task, it differs from standard classification as it exclusively learns from normal data to identify deviations indicative of anomalies.
Most existing research has focused on developing visual AD models for applications such as quality inspection and defect detection in industrial settings \cite{ref_industry1,ref_asymmetricST,ref_MVTecAD}. 
In computational pathology, specific disease conditions are often rare but hold critical clinical implications  \cite{ref_historare,ref_diverse}, posing a significant challenge. Therefore, AD holds great potential to improve clinical practice by enabling the identification of rare disease patterns with limited labeled data.

There are several limitations when applying existing methods to pathology images: 1) \textit{Computational bottleneck}: State-of-the-art AD models often rely on large backbone networks \cite{ref_cflow,ref_fastflow,ref_reverseST}, requiring substantial memory storage \cite{ref_padim,ref_patchcore}, or adopt generative models \cite{ref_anoddpm,ref_diffusionP} with long inference times, making them impractical for real-world applications. Gigapixel-sized whole slide images (WSIs)~\cite{ref_WSIs} impose even greater computational and storage demands; 2) \textit{Tissue Diversity}: Anomalies, e.g., scratches, damages, and missing components, in other domains, are often well-defined, whereas, in pathology images, anomalies are more complex and diverse, involving intricate alterations in cellular and tissue structures. Even normal tissues exhibit high structural heterogeneity~\cite{ref_diverse}, making it more challenging to differentiate between normal variations and true anomalies; 3) \textit{Generalization ability}: The quality of pathology images considerably varies due to multiple factors, including differences in institutions, imaging equipment, and staining protocols. These variations contribute to domain shifts, limiting the generalizability of existing AD methods across different domains; 4) \textit{Interpretability}: 
Most existing AD methods operate as black-boxes, providing minimal insight into their final decisions. This lack of interpretability restricts their adoption in real-world clinical settings, where transparency is crucial to support trustworthy and clinically relevant decision-making. 
 

To address these challenges, we propose Ano-NAViLa, a \textbf{N}ormal and \textbf{A}bnormal pathology knowledge-augmented \textbf{Vi}sion-\textbf{La}nguage model for \textbf{Ano}maly detection in pathology images. 
Ano-NAViLa is built by integrating a vision-language model (VLM) for feature extraction and semantic mapping with multi-layer perceptrons (MLPs) for anomaly clustering and detection. The model is designed to utilize two types of pathology knowledge to enhance its AD capabilities:
1) \textit{Data-driven knowledge}: Ano-NAViLa adopts a pathology VLM, pre-trained on large pathology image-text pair datasets, to obtain rich image and text representations, which are subsequently used to extract distinctive features between normal and abnormal tissues; 2) \textit{Expert (Pathologist-guided) knowledge}: Ano-NAViLa incorporates structured and curated pathology terms provided by experts, ensuring that the model learns clinically relevant patterns and improves interpretability in detecting and analyzing abnormal tissue structures. Notably, Ano-NAViLa utilizes both normal and abnormal pathology terms that illustrate the normal structure of tissues as well as structural and morphological alterations due to disease progression. 


The key advantages of Ano-NAViLa are three-folds:  
1) \textit{Computational efficiency}: Ano-NAViLa is designed as a lightweight architecture. The pre-trained VLM remains frozen, while only the MLP is optimized during training, substantially reducing computational overhead and supporting faster training and inference.
2) \textit{Robustness to variability in data}: Ano-NAViLa uses both normal and abnormal pathology terms, describing structural variations and alterations within each category and across the two categories. This approach is universally applicable to all pathology datasets, minimizing data-specific biases.
3) \textit{Transparency to pathologists}: Leveraging the image-text alignment, Ano-NAViLa provides textual description for both normal and abnormal images. This allows pathologists to understand, validate, and interpret the predictions, making the AD process becomes interpretable and clinically meaningful. 


To evaluate Ano-NAViLa, we employ two independent datasets for lymph node metastasis detection: 1) GastricLN: A private gastric dataset used to train, validate, and test Ano-NAViLa; 2) Camelyon16~\cite{ref_C16}: A public breast dataset used as an external test set to assess the model's generalization ability. The experimental results demonstrate that Ano-NAViLa achieves superior AD performance, surpassing existing state-of-the-art AD models. More importantly, Ano-NAViLa permits clear visualization of most metastatic regions, with improved interpretability through histologically meaningful text descriptions. In contrast, competing models struggle to distinctly localize metastases and lack interpretability. These findings highlight the effectiveness of Ano-NAViLa as an accurate, robust, and interpretable AD solution for pathology images.

\section{Related Work}
\textbf{Visual anomaly detection.} Earlier AD methods primarily adopted classical techniques like support vector machine (SVM)~\cite{ref_SVM} and principal component analysis (PCA)~\cite{ref_PCA}. Recent advances in deep learning has improved AD performance, with approaches divided into two major categories: 
1) Reconstruction-based methods, such as autoencoder \cite{ref_AE1,ref_AE3}, variational autoencoder \cite{ref_VAE2,ref_VAE3}, generative adversarial networks (GANs) \cite{ref_GAN1,ref_ganomaly,ref_GAN3}, and diffusion models \cite{ref_anoddpm,ref_diff1,ref_diffusionP}, are designed to reconstruct normal data only and utilize reconstruction errors as an indicator for AD.
Among them, diffusion models have achieved promising results but are limited by high computational cost and risk of reconstructing out-of-distribution images too well, reducing their effectiveness in AD.
2) Representation-based methods aim to learn feature representations for enhanced estimation and modeling of normal data distributions. A prominent approach is memory bank-based methods that store image features and apply various techniques, such as k-Nearest Neighbor (kNN) algorithm \cite{ref_spade,ref_patchcore} and Mahalanobis distance \cite{ref_MahalanobisAD,ref_padim}. However, these methods face scalability issues due to their large storage requirements. 
Other approaches include Gaussian distribution modeling \cite{ref_GD1,ref_GD2} for density estimation, employing normalizing flows \cite{ref_fastflow,ref_cflow,ref_csflow} to estimate the likelihood of normality, and hypersphere-based feature learning \cite{ref_patchsvdd,ref_fcdd,ref_cfa}. 

Another AD approach adopts the student–teacher framework, based on the assumption that the output difference between the two networks will be larger for anomalous data. Typically, the two networks share identical architecture \cite{ref_uninformedST,ref_stfpm,ref_efficientad}, allowing for a simple distillation process and low latency, while some explored hybrid architectures \cite{ref_multiresolutionST,ref_asymmetricST}.
VLM-based approaches have also emerged for AD. WinCLIP~\cite{ref_winclip} aggregates image features with similarities to collected normal and abnormal text prompts, while AnomalyCLIP~\cite{ref_anomalyclip} optimizes text prompts to enhance class separation. InCTRL~\cite{ref_inctrl} extracts image features guided by text prompts and detects anomalies based on residuals from few-shot normal samples. These methods leverage pretrained models for strong few/zero-shot performance, but make limited use of the interpretability offered by VLMs.

\textbf{Pathology vision-language models.} VLMs serve as a foundation for numerous vision-language tasks. Notable examples include CLIP~\cite{ref_CLIP} and ALIGN~\cite{ref_align}, which employ contrastive learning to train a dual encoder for aligning images and texts. SimVLM~\cite{ref_simvlm} adopts generative image-to-text learning with an encoder-decoder structure. CoCa~\cite{ref_coca} combines both contrastive and generative learning for improved multimodal image-text representations. Driven by these advancements, several VLMs have been introduced for pathology image analysis. For instance, PLIP~\cite{ref_PLIP} is trained on a 208K pathology images paired with natural language descriptions, BiomedCLIP~\cite{ref_biomedclip} is trained on multimodal medical dataset covering microscopy, biosignals, scientific diagrams, and flowcharts, QuiltNet~\cite{ref_quiltnet} is trained on 1M pathology image-text pairs, and CONCH~\cite{ref_CONCH} is based on CoCa and trained on a dataset with 1.17 million image-caption pairs.
\section{Method}
\begin{figure*}
\centering
\includegraphics[width=1.0\linewidth]{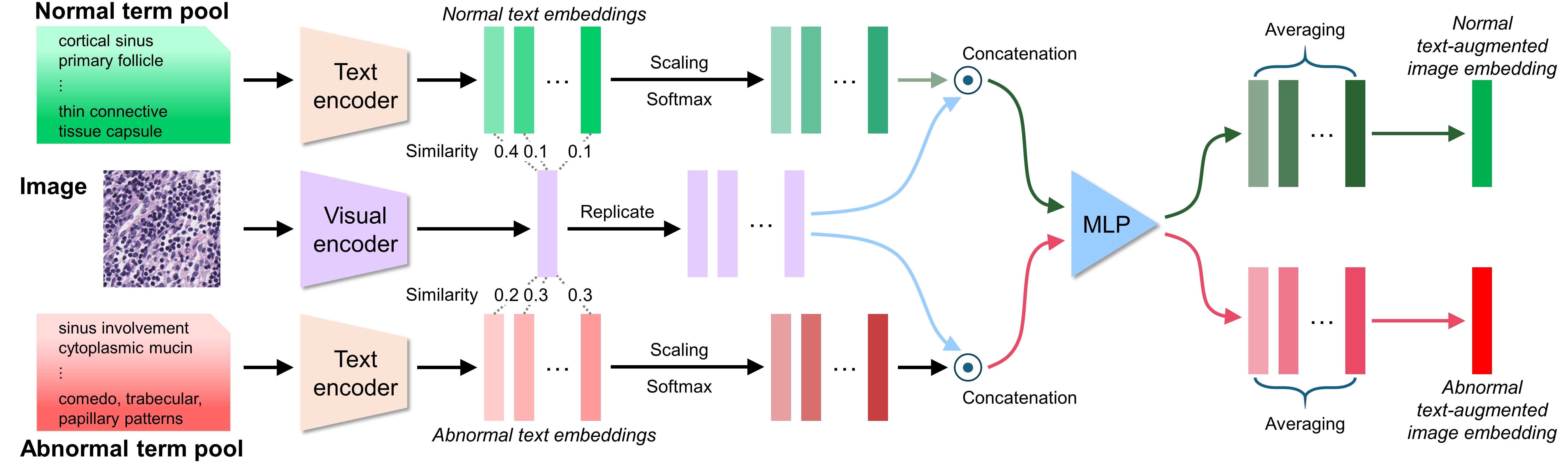}
\caption{Generation of normal and abnormal text-augmented image embeddings. Two embeddings are generated for each image, representing its relationship with the normal and abnormal term pools, respectively. Only the MLP is trainable throughout the entire process.}
\label{fig:fig1}
\end{figure*}
\begin{figure*}
\centering
\includegraphics[width=1.0\linewidth]{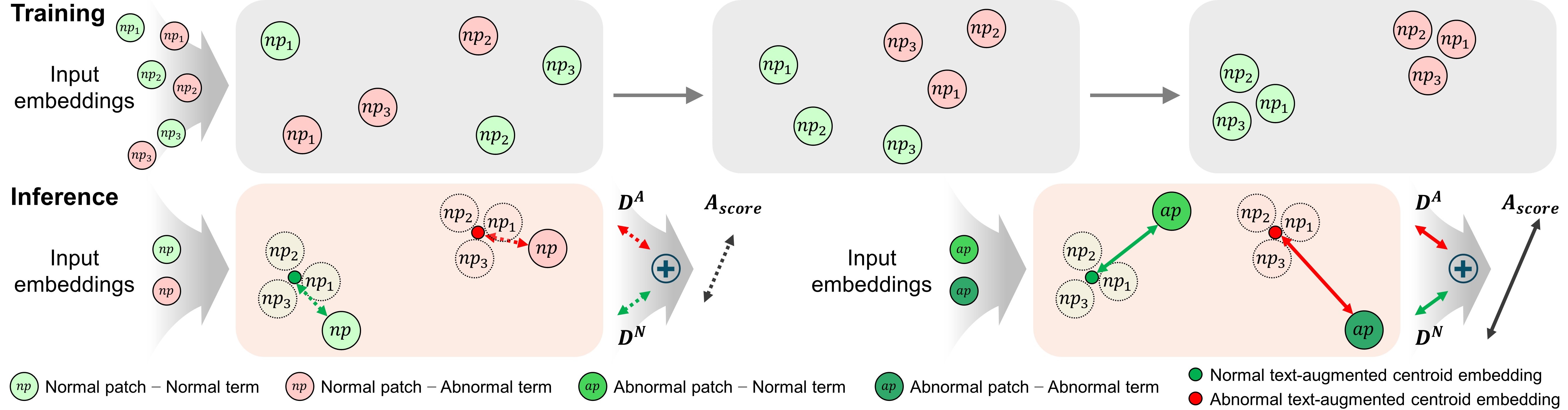}
\caption{Training and inference procedures. The MLP is trained using only normal images to separately cluster embeddings that are representative of normal image–normal term and normal image–abnormal term relations. At inference, the distances from the two cluster centroids are computed, and their sum serves as the anomaly score, with larger values indicative of anomalies.}
\label{fig:fig2}
\end{figure*}
Ano-NAViLa consists of four key components: 1) normal and abnormal term pools; 2) a VLM with visual and text encoders; 3) a trainable MLP; 4) an anomaly scoring scheme. Using a pre-trained VLM $\mathcal{V}$, an input image $x$ is processed by its visual encoder, producing an image embedding. Simultaneously, the text encoder of $\mathcal{V}$ converts terms from the two term pools into text embeddings. Through image-text alignment, these text embeddings are normalized and concatenated with the image embedding. Finally, the MLP produces text-augmented image embeddings, which are used to compute anomaly scores, effectively distinguish normal images from abnormal images.

\subsection{Construction of term pools}
We collect a set of pathology terms related to the target pathology images. Focusing on AD, these terms are categorized into two distinct pools, forming a normal term pool $T^N=\{t^N_i\}^{n_N}_{i=1}$ and an abnormal term pool $T^A=\{t^A_j\}^{n_A}_{j=1}$ where $t^N_i$ and $t^A_j$ represent the $i$th normal term and the $j$th abnormal term, respectively, and $n_N$ and $n_A$ denote the total number of terms in the normal and abnormal term pools, respectively.
The key idea is that images interact differently with the normal and abnormal terms. By training a model to recognize these distinctions, we aim to emulate human experts' decision-making process, identifying anomalous images through their associations and contrasts with both normal and abnormal pathology terms. 
For lymph node metastasis detection, relevant pathology terms are collected from literature, mostly based on~\cite{ref_terms}, thoroughly selected and validated by an experienced pathologist. The curated term pools consist of 92 normal and 48 abnormal pathology terms, which are available in the supplementary material.


\subsection{Text-augmented image representations}
Ano-NAViLa is inspired by TQx~\cite{ref_TQx}, which yields the representation of an image as a weighted sum of the top-k text embeddings from a pathology term pool. We generate VLM-based feature representations for pathology images by concatenating the image embedding with each text embedding, which are then processed by a MLP to produce dynamic representations. We use two distinct pathology term pools—normal and abnormal—to learn their specific associations with each image for AD. \cref{fig:fig1} illustrates the procedure of generating VLM-based feature representations.

Suppose that we are given a normal pathology image $x$. Let $\Gamma$ be a VLM and $\rho$ be a MLP. $\Gamma$ has an visual encoder $\Gamma^V$ and a text encoder $\Gamma^T$. For $x$, we generate an image embedding $\textbf{v}^I = \Gamma^V (x) \in \mathbb{R}^{512}$. Provided with $T^N$ and $T^A$, we produce their corresponding text embeddings $\textbf{v}^N = \{\Gamma^T(t^N_i)\}_{i=1}^{n_N}$ and $\textbf{v}^A = \{\Gamma^T(t^A_j) \}_{j=1}^{n_A}$, both in $\mathbb{R}^{512}$, respectively. 
Each text embedding is associated with the image embedding $\textbf{v}^I$ using cosine similarity ($c_{sim}$). The similarity scores are then normalized through the Softmax function and serve as weights for the text embeddings:
\begin{equation}
\scriptsize
w^{N}_i = \frac{\exp[ c_{sim} ( \mathbf{v}^{I}, \mathbf{v}^N_i ) ]}{\sum_k \exp[ c_{sim} ( \mathbf{v}^{I}, \mathbf{v}^N_k ) ]} , 
w^{A}_j = \frac{\exp[ c_{sim} ( \mathbf{v}^{I}, \mathbf{v}^A_j ) ]}{\sum_k \exp[ c_{sim} ( \mathbf{v}^{I}, \mathbf{v}^A_k ) ]}
\end{equation}
where $w^{N}_i$ and $w^{A}_j$ are the corresponding weights for $\mathbf{v}^N_i$ and $\mathbf{v}^A_j$, respectively. These weights, which typically fall within the range $[0, 0.1]$, are scale up using an exponential function. After weighing the text embeddings, each text embedding is concatenated with the image embedding $\mathbf{v}^I$, forming a vector in $\mathbb{R}^{1024}$, which is then passed through the MLP $\rho$, yielding two image-text embeddings as follows:
\begin{equation}
\footnotesize
\mathbf{u}^{N}_i = \rho \bigl( \mathbf{v}^I \circ (e^{w^N_i} \cdot \mathbf{v}^N_i) \bigr), 
\mathbf{u}^{A}_j = \rho \bigl( \mathbf{v}^I \circ (e^{w^A_j} \cdot \mathbf{v}^A_j) \bigr)
\end{equation}
where $\mathbf{u}^{N}_i$ and $\mathbf{u}^{A}_j$ denote the $i$th image-normal and $j$th image-abnormal text embeddings, respectively, and $\circ$ is the concatenation operation. Finally, by averaging all image-text embeddings in each category, we obtain the representative normal text-augmented image embedding $\mathbf{h}^N$ and abnormal text-augmented image embedding $\mathbf{h}^A$ for the given image $x$. 
In this manner, pathology terms with higher similarity with the image $x$ contribute more to the final representations. Consequently, the final representation reflects the relationship between the image and the corresponding text pool.
We note that we employ CONCH~\cite{ref_CONCH} as $\Gamma$, which is frozen throughout the entire procedure, and the MLP $\rho$ is the only trainable component.

\subsection{Optimization: Contrastive learning}
The goal of the MLP $\rho$ is to obtain text-augmented image embeddings in a latent space where embeddings within the same category (i.e., driven by normal pathology terms or abnormal pathology terms) are closely clustered, while two categories are pushed further apart to enhance separability. 
To achieve this, we propose the following loss function:
\begin{equation}
\footnotesize
\mathcal{L} = -\log\left[\frac{S_{\text{intra}}(\mathbf{h}^N) + S_{\text{intra}}(\mathbf{h}^A)}{S_{\text{intra}}(\mathbf{h}^N) + S_{\text{intra}}(\mathbf{h}^A) + S_{\text{inter}}(\mathbf{h}^N, \mathbf{h}^A)} \right],
\end{equation}
\begingroup
\footnotesize
\begin{gather}
{S}_{\text{intra}}(\mathbf{h}^N) = \frac{1}{Z_1} \sum_{i=1}^{B-1}\sum_{j=i+1}^{B}\mathrm{exp}[{{c_{sim}}(\mathbf{h}^{N}_{i}, \mathbf{h}^{N}_{j})}], \\
{S}_{\text{intra}}(\mathbf{h}^A) = \frac{1}{Z_1} \sum_{i=1}^{B-1}\sum_{j=i+1}^{B}\mathrm{exp}[{{c_{sim}}(\mathbf{h}^{A}_{i}, \mathbf{h}^{A}_{j})}], \\
{S}_{\text{inter}}(\mathbf{h}^N, \mathbf{h}^A) = \frac{1}{Z_2} \sum_{i=1}^{B}\sum_{j=1}^{B}\mathrm{exp}[{{c_{sim}}(\mathbf{h}^{N}_{i}, \mathbf{h}^{A}_{j})}]
\end{gather}
\endgroup
\noindent where $B$ is the batch size, $\mathbf{h}^N_i$ and $\mathbf{h}^A_i$ denote the normal and abnormal text-augmented image embeddings for the $i$th image $x_i$, respectively, and $Z_1 = \binom{B}{2}$ and $Z_2 = B^2$ are normalization factors. 
By minimizing $\mathcal{L}$, we aim to maximize intra-group similarity ${S}_{\text{intra}}$ for each pool category, while minimizing inter-group similarity ${S}_{\text{inter}}$ between the two pool categories. 
\cref{fig:fig2} illustrates the training process.
Though the MLP $\rho$ is trained exclusively on normal images without involving abnormal images during optimization, the interaction between normal images with abnormal pathology terms function as a regularization mechanism, further enhancing representation power of normal images.

\subsection{Anomaly score and inference}
By training the MLP, the model learns two distinct relationships: normal images to normal terms, and normal images to abnormal terms. The two resulting clusters, representing these relationships, are located separately in the feature space. Therefore, an abnormal image is expected to align differently with both normal and abnormal terms compared to normal images, which allows us to compute an anomaly score based on its relative position to the two clusters.

To facilitate AD at the patch-level, we first compute two centroid text-augmented image embeddings from normal images, denoted as ${\bar{\mathbf{h}}}^{N}$ and ${\bar{\mathbf{h}}}^{A}$, for the normal and abnormal pathology term pools, respectively. These two centroid embeddings are used to calculate the deviations from the normal distribution. Specifically, given an input image $x$, we produce the corresponding embeddings $\mathbf{h}^N$ and $\mathbf{h}^A$, and then compute two deviation scores as follows:
\begin{center}
\begingroup
\footnotesize
\vspace{-1em}
\begin{minipage}{0.9\linewidth}
\begin{gather}
{D^{N}}(\mathbf{h}^N) = 1 - {c_{sim}}(\mathbf{h}^N, {\bar{\mathbf{h}}}^{N}) \in [0, 2],  \\
{D^{A}}(\mathbf{h}^A) = 1 - {c_{sim}}(\mathbf{h}^A, {\bar{\mathbf{h}}}^{A}) \in [0, 2] 
\end{gather}
\end{minipage}
\endgroup
\end{center}
\noindent These two scores are combined to produce the patch-level anomaly score as ${A}_{score} = D^{N}(\mathbf{h}^N) + D^{A}(\mathbf{h}^A)$. 

At inference, given a normal input image, its associations with both normal and abnormal pathology terms should closely resemble those observed during training. In contrast, an abnormal image is likely to exhibit different interactions with both term pools, resulting in higher deviation scores. Consequently, abnormal images tend to produce higher anomaly scores compared to normal images.

For the WSI-level AD, splitting a WSI into a set of disjoint image patches, we compute the patch-level anomaly score $A_{score}$ for each patch and assign the scores to their corresponding locations, thereby creating an anomaly score heatmap. Following~\cite{ref_diffusionP}, we apply 3x3 erosion operation to smooth the heatmap.
Two different approaches are employed to obtain the final WSI-level anomaly score: 1) Maximum anomaly score: ${A}^{\max}_{score} = max (A_{score})$ and 2) Average of the top 1\% anomaly scores:  $A^{\text{top1\%}}_{score} = \frac{1}{|\mathcal{K}|} \sum_{k \in \mathcal{K}} A_{score} (k)$ where $\mathcal{K}$ represents top 1\% patches.


\section{Experiments and Results}

\textbf{Dataset.} Two lymph node datasets are utilized in this study: \textbf{GastricLN} and \textbf{Camelyon16}~\cite{ref_C16}. \textbf{GastricLN} is composed of 808 gastric lymph node WSIs (751 normal and 57 metastasis), digitized at 20x magnification, acquired from two hospitals. These WSIs are divided into training (643 normal WSIs), validation (50 normal WSIs), and test (58 normal and 57 metastasis WSIs) sets.
\textbf{Camelyon16} is solely used for external testing, comprising 129 WSIs of breast lymph node, scanned at 40x magnification, (80 normal WSIs and 49 metastasis WSIs). Following ~\cite{ref_diffusionP}, 22 macro-metastasis WSIs with $\geq$ 2mm tumor cell diameter and 80 normal WSIs form \textbf{Camelyon16$_{\textbf{macro}}$} and are used to assess the effect of metastatic regions size on AD performance. Since most WSIs in \textbf{GastricLN} contain macro-metastases, we do not explicitly define a macro dataset for \textbf{GastricLN}.
Following~\cite{ref_patchextraction}, WSIs are split into patches of size 256×256 pixels and 512×512 pixels for \textbf{GastricLN} and \textbf{Camelyon16}, respectively.

\noindent\textbf{Evaluation metric.} 
To evaluate AD performance at patch-level, we employ the area under the receiver operating characteristic curve (AUROC). For WSI-level performance, both AUROC and the area under the precision-recall curve (AUPR) are utilized. We compute the 95\% confidence interval for both AUROC and AUPR by 2000-fold bootstrapping by 100\% resampling with replacement.

\noindent\textbf{Comparative experiments.}
Seven visual AD models are employed for comparison, including two student-teacher framework-based methods: STFPM~\cite{ref_stfpm} and EfficientAD~\cite{ref_efficientad}, two representation-based methods: FastFlow~\cite{ref_fastflow} and CFA~\cite{ref_cfa}, two reconstruction-based methods: GANomaly~\cite{ref_ganomaly} and AnoDDPM~\cite{ref_anoddpm}, and one VLM-based method: AnomalyCLIP~\cite{ref_anomalyclip}. The details of these methods are illustrated in the supplementary material.

\noindent\textbf{Implementation detail.} We use the text prompt \texttt{"an image showing KEYWORD"} for the text encoder of VLM $\Gamma^T$, where \texttt{KEYWORD} is replaced by pathology terms. The MLP consists of three linear layers with ReLU activations in the first two, mapping a 1024-dimensional input to a 128-dimensional output. For optimization, we employ the Adam optimizer with a learning rate of 0.001 and a batch size of 100. For each batch, gradients are accumulated and updated after processing 100 batches. The training process runs for a single epoch on the training set of \textbf{GastricLN} due to its large size. The normal and abnormal text-augmented centroid embeddings are obtained from the validation set of \textbf{GastricLN}.  
For computational reasons, AnoDDPM is evaluated on an NVIDIA RTX A6000, while the other models are evaluated on an NVIDIA RTX 3090 GPU. The specific training strategies for the seven competing models are illustrated in the supplementary material.
\begin{table*}[]
\centering
\resizebox{\textwidth}{!}{%
\begin{tabular}{|l|cccc|c|cccc|cccc|}
\hline
\multirow{2}{*}{Method} &
  \multicolumn{4}{c|}{GastricLN (WSI)} &
  Patch &
  \multicolumn{4}{c|}{Camelyon16} &
  \multicolumn{4}{c|}{Camelyon16$_{\text{macro}}$} \\ \cline{2-14} 
 &
  \begin{tabular}[c]{@{}c@{}}AUROC\\ ($A^{\max}_{score}$)\end{tabular} &
  \begin{tabular}[c]{@{}c@{}}AUPR\\ ($A^{\max}_{score}$)\end{tabular} &
  \begin{tabular}[c]{@{}c@{}}AUROC\\ ($A^{\text{top1\%}}_{score}$)\end{tabular} &
  \begin{tabular}[c]{@{}c@{}}AUPR\\ ($A^{\text{top1\%}}_{score}$)\end{tabular} &
  AUROC &
  \begin{tabular}[c]{@{}c@{}}AUROC\\ ($A^{\max}_{score}$)\end{tabular} &
  \begin{tabular}[c]{@{}c@{}}AUPR\\ ($A^{\max}_{score}$)\end{tabular} &
  \begin{tabular}[c]{@{}c@{}}AUROC\\ ($A^{\text{top1\%}}_{score}$)\end{tabular} &
  \begin{tabular}[c]{@{}c@{}}AUPR\\ ($A^{\text{top1\%}}_{score}$)\end{tabular} &
  \begin{tabular}[c]{@{}c@{}}AUROC\\ ($A^{\max}_{score}$)\end{tabular} &
  \begin{tabular}[c]{@{}c@{}}AUPR\\ ($A^{\max}_{score}$)\end{tabular} &
  \begin{tabular}[c]{@{}c@{}}AUROC\\ ($A^{\text{top1\%}}_{score}$)\end{tabular} &
  \begin{tabular}[c]{@{}c@{}}AUPR\\ ($A^{\text{top1\%}}_{score}$)\end{tabular} \\ \hline
GANomaly &
  \begin{tabular}[c]{@{}c@{}}0.3947\\[-1.6mm] {\scriptsize {[}0.30, 0.50{]}}\end{tabular} &
  \begin{tabular}[c]{@{}c@{}}0.4323\\[-1.6mm] {\scriptsize {[}0.33, 0.55{]}}\end{tabular} &
  \begin{tabular}[c]{@{}c@{}}0.3125\\[-1.6mm] {\scriptsize {[}0.22, 0.41{]}}\end{tabular} &
  \begin{tabular}[c]{@{}c@{}}0.3958\\[-1.6mm] {\scriptsize {[}0.30, 0.50{]}}\end{tabular} &
  \begin{tabular}[c]{@{}c@{}}0.4182\\[-1.6mm] {\scriptsize {[}0.42, 0.42{]}}\end{tabular} &
  \begin{tabular}[c]{@{}c@{}}0.6503\\[-1.6mm] {\scriptsize {[}0.55, 0.75{]}}\end{tabular} &
  \begin{tabular}[c]{@{}c@{}}0.5519\\[-1.6mm] {\scriptsize {[}0.41, 0.70{]}}\end{tabular} &
  \begin{tabular}[c]{@{}c@{}}0.6643\\[-1.6mm] {\scriptsize {[}0.57, 0.76{]}}\end{tabular} &
  \begin{tabular}[c]{@{}c@{}}0.5445\\[-1.6mm] {\scriptsize {[}0.41, 0.69{]}}\end{tabular} &
  \begin{tabular}[c]{@{}c@{}}0.6631\\[-1.6mm] {\scriptsize {[}0.54, 0.79{]}}\end{tabular} &
  \begin{tabular}[c]{@{}c@{}}0.3873\\[-1.6mm] {\scriptsize {[}0.21, 0.57{]}}\end{tabular} &
  \begin{tabular}[c]{@{}c@{}}0.6494\\[-1.6mm] {\scriptsize {[}0.51, 0.78{]}}\end{tabular} &
  \begin{tabular}[c]{@{}c@{}}0.3575\\[-1.6mm] {\scriptsize {[}0.19, 0.55{]}}\end{tabular} \\
STFPM &
  \begin{tabular}[c]{@{}c@{}}0.9779\\[-1.6mm] {\scriptsize {[}0.95, 1.00{]}}\end{tabular} &
  \begin{tabular}[c]{@{}c@{}}0.9728\\[-1.6mm] {\scriptsize {[}0.93, 1.00{]}}\end{tabular} &
  \begin{tabular}[c]{@{}c@{}}0.9797\\[-1.6mm] {\scriptsize {[}0.95, 1.00{]}}\end{tabular} &
  \begin{tabular}[c]{@{}c@{}}0.9714\\[-1.6mm] {\scriptsize {[}0.93, 1.00{]}}\end{tabular} &
  \begin{tabular}[c]{@{}c@{}}0.9538\\[-1.6mm] {\scriptsize {[}0.95, 0.96{]}}\end{tabular} &
  \begin{tabular}[c]{@{}c@{}}0.7158\\[-1.6mm] {\scriptsize {[}0.62, 0.80{]}}\end{tabular} &
  \begin{tabular}[c]{@{}c@{}}0.6862\\[-1.6mm] {\scriptsize {[}0.55, 0.79{]}}\end{tabular} &
  \begin{tabular}[c]{@{}c@{}}0.7324\\[-1.6mm] {\scriptsize {[}0.64, 0.82{]}}\end{tabular} &
  \begin{tabular}[c]{@{}c@{}}0.7038\\[-1.6mm] {\scriptsize {[}0.58, 0.81{]}}\end{tabular} &
  \begin{tabular}[c]{@{}c@{}}0.7710\\[-1.6mm] {\scriptsize {[}0.63, 0.89{]}}\end{tabular} &
  \begin{tabular}[c]{@{}c@{}}0.6314\\[-1.6mm] {\scriptsize {[}0.42, 0.80{]}}\end{tabular} &
  \begin{tabular}[c]{@{}c@{}}0.7818\\[-1.6mm] {\scriptsize {[}0.65, 0.90{]}}\end{tabular} &
  \begin{tabular}[c]{@{}c@{}}0.6457\\[-1.6mm] {\scriptsize {[}0.42, 0.82{]}}\end{tabular} \\
FastFlow &
  \begin{tabular}[c]{@{}c@{}}0.9301\\[-1.6mm] {\scriptsize {[}0.87, 0.98{]}}\end{tabular} &
  \begin{tabular}[c]{@{}c@{}}0.8898\\[-1.6mm] {\scriptsize {[}0.78, 0.97{]}}\end{tabular} &
  \begin{tabular}[c]{@{}c@{}}0.9446\\[-1.6mm] {\scriptsize {[}0.89, 1.00{]}}\end{tabular} &
  \begin{tabular}[c]{@{}c@{}}0.8761\\[-1.6mm] {\scriptsize {[}0.76, 0.99{]}}\end{tabular} &
  \begin{tabular}[c]{@{}c@{}}0.9242\\[-1.6mm] {\scriptsize {[}0.92, 0.93{]}}\end{tabular} &
  \begin{tabular}[c]{@{}c@{}}0.6990\\[-1.6mm] {\scriptsize {[}0.61, 0.79{]}}\end{tabular} &
  \begin{tabular}[c]{@{}c@{}}0.6304\\[-1.6mm] {\scriptsize {[}0.50, 0.75{]}}\end{tabular} &
  \begin{tabular}[c]{@{}c@{}}0.7337\\[-1.6mm] {\scriptsize {[}0.64, 0.82{]}}\end{tabular} &
  \begin{tabular}[c]{@{}c@{}}0.6335\\[-1.6mm] {\scriptsize {[}0.48, 0.78{]}}\end{tabular} &
  \begin{tabular}[c]{@{}c@{}}0.7239\\[-1.6mm] {\scriptsize {[}0.59, 0.85{]}}\end{tabular} &
  \begin{tabular}[c]{@{}c@{}}0.5291\\[-1.6mm] {\scriptsize {[}0.31, 0.71{]}}\end{tabular} &
  \begin{tabular}[c]{@{}c@{}}0.7705\\[-1.6mm] {\scriptsize {[}0.64, 0.88{]}}\end{tabular} &
  \begin{tabular}[c]{@{}c@{}}0.4976\\[-1.6mm] {\scriptsize {[}0.30, 0.75{]}}\end{tabular} \\
CFA &
  \begin{tabular}[c]{@{}c@{}}0.9673\\[-1.6mm] {\scriptsize {[}0.93, 1.00{]}}\end{tabular} &
  \begin{tabular}[c]{@{}c@{}}0.9243\\[-1.6mm] {\scriptsize {[}0.82, 1.00{]}}\end{tabular} &
  \begin{tabular}[c]{@{}c@{}}0.9570\\[-1.6mm] {\scriptsize {[}0.90, 1.00{]}}\end{tabular} &
  \begin{tabular}[c]{@{}c@{}}0.8876\\[-1.6mm] {\scriptsize {[}0.77, 1.00{]}}\end{tabular} &
  \begin{tabular}[c]{@{}c@{}}0.8881\\[-1.6mm] {\scriptsize {[}0.89, 0.89{]}}\end{tabular} &
  \begin{tabular}[c]{@{}c@{}}0.7003\\[-1.6mm] {\scriptsize {[}0.60, 0.79{]}}\end{tabular} &
  \begin{tabular}[c]{@{}c@{}}0.6435\\[-1.6mm] {\scriptsize {[}0.50, 0.76{]}}\end{tabular} &
  \begin{tabular}[c]{@{}c@{}}0.7355\\[-1.6mm] {\scriptsize {[}0.64, 0.83{]}}\end{tabular} &
  \begin{tabular}[c]{@{}c@{}}0.6718\\[-1.6mm] {\scriptsize {[}0.53, 0.80{]}}\end{tabular} &
  \begin{tabular}[c]{@{}c@{}}0.7574\\[-1.6mm] {\scriptsize {[}0.61, 0.88{]}}\end{tabular} &
  \begin{tabular}[c]{@{}c@{}}0.6008\\[-1.6mm] {\scriptsize {[}0.37, 0.79{]}}\end{tabular} &
  \begin{tabular}[c]{@{}c@{}}0.7699\\[-1.6mm] {\scriptsize {[}0.63, 0.90{]}}\end{tabular} &
  \begin{tabular}[c]{@{}c@{}}0.5557\\[-1.6mm] {\scriptsize {[}0.34, 0.78{]}}\end{tabular} \\
AnoDDPM &
  \begin{tabular}[c]{@{}c@{}}0.8860\\[-1.6mm] {\scriptsize {[}0.82, 0.94{]}}\end{tabular} &
  \begin{tabular}[c]{@{}c@{}}0.8739\\[-1.6mm] {\scriptsize {[}0.78, 0.94{]}}\end{tabular} &
  \begin{tabular}[c]{@{}c@{}}0.8621\\[-1.6mm] {\scriptsize {[}0.79, 0.93{]}}\end{tabular} &
  \begin{tabular}[c]{@{}c@{}}0.8403\\[-1.6mm] {\scriptsize {[}0.73, 0.93{]}}\end{tabular} &
  \begin{tabular}[c]{@{}c@{}}0.8860\\[-1.6mm] {\scriptsize {[}0.88, 0.89{]}}\end{tabular} &
  \begin{tabular}[c]{@{}c@{}}0.4816\\[-1.6mm] {\scriptsize {[}0.37, 0.63{]}}\end{tabular} &
  \begin{tabular}[c]{@{}c@{}}0.4012\\[-1.6mm] {\scriptsize {[}0.15, 0.40{]}}\end{tabular} &
  \begin{tabular}[c]{@{}c@{}}0.5013\\[-1.6mm] {\scriptsize {[}0.34, 0.64{]}}\end{tabular} &
  \begin{tabular}[c]{@{}c@{}}0.4471\\[-1.6mm] {\scriptsize {[}0.17, 0.49{]}}\end{tabular} &
  \begin{tabular}[c]{@{}c@{}}0.5386\\[-1.6mm] {\scriptsize {[}0.42, 0.66{]}}\end{tabular} &
  \begin{tabular}[c]{@{}c@{}}0.2259\\[-1.6mm] {\scriptsize {[}0.15, 0.36{]}}\end{tabular} &
  \begin{tabular}[c]{@{}c@{}}0.6631\\[-1.6mm] {\scriptsize {[}0.54, 0.78{]}}\end{tabular} &
  \begin{tabular}[c]{@{}c@{}}0.4353\\[-1.6mm] {\scriptsize {[}0.25, 0.62{]}}\end{tabular} \\
EfficientAD &
  \begin{tabular}[c]{@{}c@{}}0.8772\\[-1.6mm] {\scriptsize {[}0.79, 0.95{]}}\end{tabular} &
  \begin{tabular}[c]{@{}c@{}}0.7856\\[-1.6mm] {\scriptsize {[}0.65, 0.91{]}}\end{tabular} &
  \begin{tabular}[c]{@{}c@{}}0.8817\\[-1.6mm] {\scriptsize {[}0.80, 0.95{]}}\end{tabular} &
  \begin{tabular}[c]{@{}c@{}}0.7539\\[-1.6mm] {\scriptsize {[}0.62, 0.91{]}}\end{tabular} &
  \begin{tabular}[c]{@{}c@{}}0.8432\\[-1.6mm] {\scriptsize {[}0.84, 0.85{]}}\end{tabular} &
  \begin{tabular}[c]{@{}c@{}}0.6462\\[-1.6mm] {\scriptsize {[}0.55, 0.74{]}}\end{tabular} &
  \begin{tabular}[c]{@{}c@{}}0.5170\\[-1.6mm] {\scriptsize {[}0.38, 0.66{]}}\end{tabular} &
  \begin{tabular}[c]{@{}c@{}}0.6526\\[-1.6mm] {\scriptsize {[}0.55, 0.75{]}}\end{tabular} &
  \begin{tabular}[c]{@{}c@{}}0.5364\\[-1.6mm] {\scriptsize {[}0.40, 0.69{]}}\end{tabular} &
  \begin{tabular}[c]{@{}c@{}}0.6415\\[-1.6mm] {\scriptsize {[}0.51, 0.77{]}}\end{tabular} &
  \begin{tabular}[c]{@{}c@{}}0.3067\\[-1.6mm] {\scriptsize {[}0.17, 0.51{]}}\end{tabular} &
  \begin{tabular}[c]{@{}c@{}}0.6006\\[-1.6mm] {\scriptsize {[}0.45, 0.74{]}}\end{tabular} &
  \begin{tabular}[c]{@{}c@{}}0.2928\\[-1.6mm] {\scriptsize {[}0.16, 0.48{]}}\end{tabular} \\
AnomalyCLIP &
  \begin{tabular}[c]{@{}c@{}}0.7371\\[-1.6mm] {\scriptsize {[}0.64, 0.82{]}}\end{tabular} &
  \begin{tabular}[c]{@{}c@{}}0.7057\\[-1.6mm] {\scriptsize {[}0.58, 0.83{]}}\end{tabular} &
  \begin{tabular}[c]{@{}c@{}}0.5309\\[-1.6mm] {\scriptsize {[}0.43, 0.64{]}}\end{tabular} &
  \begin{tabular}[c]{@{}c@{}}0.5054\\[-1.6mm] {\scriptsize {[}0.40, 0.65{]}}\end{tabular} &
  \begin{tabular}[c]{@{}c@{}}0.3314\\[-1.6mm] {\scriptsize {[}0.33, 0.33{]}}\end{tabular} &
  \begin{tabular}[c]{@{}c@{}}0.6003\\[-1.6mm] {\scriptsize {[}0.50, 0.69{]}}\end{tabular} &
  \begin{tabular}[c]{@{}c@{}}0.4830\\[-1.6mm] {\scriptsize {[}0.35, 0.62{]}}\end{tabular} &
  \begin{tabular}[c]{@{}c@{}}0.5446\\[-1.6mm] {\scriptsize {[}0.44, 0.65{]}}\end{tabular} &
  \begin{tabular}[c]{@{}c@{}}0.4568\\[-1.6mm] {\scriptsize {[}0.33, 0.58{]}}\end{tabular} &
  \begin{tabular}[c]{@{}c@{}}0.6187\\[-1.6mm] {\scriptsize {[}0.48, 0.75{]}}\end{tabular} &
  \begin{tabular}[c]{@{}c@{}}0.3480\\[-1.6mm] {\scriptsize {[}0.18, 0.54{]}}\end{tabular} &
  \begin{tabular}[c]{@{}c@{}}0.5392\\[-1.6mm] {\scriptsize {[}0.39, 0.69{]}}\end{tabular} &
  \begin{tabular}[c]{@{}c@{}}0.3402\\[-1.6mm] {\scriptsize {[}0.17, 0.52{]}}\end{tabular} \\ \hline
Ano-NAViLa &
  \textbf{\begin{tabular}[c]{@{}c@{}}0.9967\\[-1.6mm] {\scriptsize {[}0.99, 1.00{]}}\end{tabular}} &
  \textbf{\begin{tabular}[c]{@{}c@{}}0.9971\\[-1.6mm] {\scriptsize {[}0.99, 1.00{]}}\end{tabular}} &
  \textbf{\begin{tabular}[c]{@{}c@{}}0.9894\\[-1.6mm] {\scriptsize {[}0.97, 1.00{]}}\end{tabular}} &
  \textbf{\begin{tabular}[c]{@{}c@{}}0.9904\\[-1.6mm] {\scriptsize {[}0.98, 1.00{]}}\end{tabular}} &
  \textbf{\begin{tabular}[c]{@{}c@{}}0.9681\\[-1.6mm] {\scriptsize {[}0.97, 0.97{]}}\end{tabular}} &
  \textbf{\begin{tabular}[c]{@{}c@{}}0.8594\\[-1.6mm] {\scriptsize {[}0.79, 0.92{]}}\end{tabular}} &
  \textbf{\begin{tabular}[c]{@{}c@{}}0.8309\\[-1.6mm] {\scriptsize {[}0.73, 0.91{]}}\end{tabular}} &
  \textbf{\begin{tabular}[c]{@{}c@{}}0.7702\\[-1.6mm] {\scriptsize {[}0.67, 0.86{]}}\end{tabular}} &
  \textbf{\begin{tabular}[c]{@{}c@{}}0.7941\\[-1.6mm] {\scriptsize {[}0.69, 0.88{]}}\end{tabular}} &
  \textbf{\begin{tabular}[c]{@{}c@{}}0.9858\\[-1.6mm] {\scriptsize {[}0.96, 1.00{]}}\end{tabular}} &
  \textbf{\begin{tabular}[c]{@{}c@{}}0.9547\\[-1.6mm] {\scriptsize {[}0.88, 1.00{]}}\end{tabular}} &
  \textbf{\begin{tabular}[c]{@{}c@{}}0.9761\\[-1.6mm] {\scriptsize {[}0.92, 1.00{]}}\end{tabular}} &
  \textbf{\begin{tabular}[c]{@{}c@{}}0.9699\\[-1.6mm] {\scriptsize {[}0.89, 1.00{]}}\end{tabular}} \\ \hline
\end{tabular}
}
\caption{AD performance on GastricLN and Camelyon16 datasets.}
\label{tab:tab1}
\end{table*}
\begin{table*}[htbp]
\centering
\begin{minipage}[b]{0.3\textwidth}
    \centering
    \renewcommand{\arraystretch}{1.56}
    \resizebox{\textwidth}{!}{%
\begin{tabular}{|l|ccc|}
\hline
Method &
  \begin{tabular}[c]{@{}c@{}}Params\\ {[}M{]}\end{tabular} &
  \begin{tabular}[c]{@{}c@{}}Latency\\ {[}ms{]}\end{tabular} &
  \begin{tabular}[c]{@{}c@{}}GPU\\ {[}MiB{]}\end{tabular} \\ \hline
GANomaly    & 188.69    & \textbf{1.45}  & 1580 \\
STFPM       & 2.78      & 14.76           & \textbf{504} \\
FastFlow    & 45.01     & 26.21           & 1700 \\
CFA         & 31.29     & 8.58           & 10990 \\
AnoDDPM     & 113.67      & 915.56           & 10712 \\
EfficientAD & 8.06      & 3.86           & 2774 \\ 
AnomalyCLIP & 5.56      & 16.86           & 3176 \\ \hline
Ano-NAViLa   & \textbf{0.69} & 3.61       & 2098 \\ \hline
\end{tabular}
    }
\caption{Computational efficiency.}
\label{tab:tab2}
\end{minipage}%
\hspace{0pt}
\begin{minipage}[b]{0.69\textwidth}
    \centering
    \resizebox{\textwidth}{!}{%
\begin{tabular}{|l|cccc|c|cccc|}
\hline
\multirow{2}{*}{Method} &
  \multicolumn{4}{c|}{GastricLN (WSI)} &
  Patch &
  \multicolumn{4}{c|}{Camelyon16} \\ \cline{2-10} 
 &
  \begin{tabular}[c]{@{}c@{}}AUROC\\ ($A^{\max}_{score}$)\end{tabular} &
  \begin{tabular}[c]{@{}c@{}}AUPR\\ ($A^{\max}_{score}$)\end{tabular} &
  \begin{tabular}[c]{@{}c@{}}AUROC\\ ($A^{\text{top1\%}}_{score}$)\end{tabular} &
  \begin{tabular}[c]{@{}c@{}}AUPR\\ ($A^{\text{top1\%}}_{score}$)\end{tabular} &
  AUROC &
  \begin{tabular}[c]{@{}c@{}}AUROC\\ ($A^{\max}_{score}$)\end{tabular} &
  \begin{tabular}[c]{@{}c@{}}AUPR\\ ($A^{\max}_{score}$)\end{tabular} &
  \begin{tabular}[c]{@{}c@{}}AUROC\\ ($A^{\text{top1\%}}_{score}$)\end{tabular} &
  \begin{tabular}[c]{@{}c@{}}AUPR\\ ($A^{\text{top1\%}}_{score}$)\end{tabular} \\ \hline
Sum &
  \begin{tabular}[c]{@{}c@{}}0.9967\\[-1.6mm] {\scriptsize {[}0.99, 1.00{]}}\end{tabular} &
  \begin{tabular}[c]{@{}c@{}}0.9971\\[-1.6mm] {\scriptsize {[}0.99, 1.00{]}}\end{tabular} &
  \begin{tabular}[c]{@{}c@{}}0.9894\\[-1.6mm] {\scriptsize {[}0.97, 1.00{]}}\end{tabular} &
  \begin{tabular}[c]{@{}c@{}}0.9904\\[-1.6mm] {\scriptsize {[}0.98, 1.00{]}}\end{tabular} &
  \textbf{\begin{tabular}[c]{@{}c@{}}0.9681\\[-1.6mm] {\scriptsize {[}0.97, 0.97{]}}\end{tabular}} &
  \begin{tabular}[c]{@{}c@{}}0.8594\\[-1.6mm] {\scriptsize {[}0.79, 0.93{]}}\end{tabular} &
  \textbf{\begin{tabular}[c]{@{}c@{}}0.8309\\[-1.6mm] {\scriptsize {[}0.73, 0.91{]}}\end{tabular}} &
  \begin{tabular}[c]{@{}c@{}}0.7702\\[-1.6mm] {\scriptsize {[}0.67, 0.86{]}}\end{tabular} &
  \begin{tabular}[c]{@{}c@{}}0.7941\\[-1.6mm] {\scriptsize {[}0.69, 0.88{]}}\end{tabular} \\
Max &
  \begin{tabular}[c]{@{}c@{}}0.9967\\[-1.6mm] {\scriptsize {[}0.99, 1.00{]}}\end{tabular} &
  \begin{tabular}[c]{@{}c@{}}0.9971\\[-1.6mm] {\scriptsize {[}0.99, 1.00{]}}\end{tabular} &
  \begin{tabular}[c]{@{}c@{}}0.9888\\[-1.6mm] {\scriptsize {[}0.97, 1.00{]}}\end{tabular} &
  \begin{tabular}[c]{@{}c@{}}0.9899\\[-1.6mm] {\scriptsize {[}0.97, 1.00{]}}\end{tabular} &
  \begin{tabular}[c]{@{}c@{}}0.9669\\[-1.6mm] {\scriptsize {[}0.97, 0.97{]}}\end{tabular} &
  \begin{tabular}[c]{@{}c@{}}0.8566\\[-1.6mm] {\scriptsize {[}0.78, 0.92{]}}\end{tabular} &
  \begin{tabular}[c]{@{}c@{}}0.8197\\[-1.6mm] {\scriptsize {[}0.71, 0.90{]}}\end{tabular} &
  \begin{tabular}[c]{@{}c@{}}0.7719\\[-1.6mm] {\scriptsize {[}0.68, 0.86{]}}\end{tabular} &
  \begin{tabular}[c]{@{}c@{}}0.7854\\[-1.6mm] {\scriptsize {[}0.68, 0.87{]}}\end{tabular} \\
L2-norm &
  \begin{tabular}[c]{@{}c@{}}0.9967\\[-1.6mm] {\scriptsize {[}0.99, 1.00{]}}\end{tabular} &
  \begin{tabular}[c]{@{}c@{}}0.9971\\[-1.6mm] {\scriptsize {[}0.99, 1.00{]}}\end{tabular} &
  \begin{tabular}[c]{@{}c@{}}0.9888\\[-1.6mm] {\scriptsize {[}0.97, 1.00{]}}\end{tabular} &
  \begin{tabular}[c]{@{}c@{}}0.9899\\[-1.6mm] {\scriptsize {[}0.97, 1.00{]}}\end{tabular} &
  \begin{tabular}[c]{@{}c@{}}0.9678\\[-1.6mm] {\scriptsize {[}0.97, 0.97{]}}\end{tabular} &
  \begin{tabular}[c]{@{}c@{}}0.8592\\[-1.6mm] {\scriptsize {[}0.79, 0.92{]}}\end{tabular} &
  \begin{tabular}[c]{@{}c@{}}0.8258\\[-1.6mm] {\scriptsize {[}0.72, 0.90{]}}\end{tabular} &
  \begin{tabular}[c]{@{}c@{}}0.7712\\[-1.6mm] {\scriptsize {[}0.68, 0.86{]}}\end{tabular} &
  \begin{tabular}[c]{@{}c@{}}0.7880\\[-1.6mm] {\scriptsize {[}0.68, 0.87{]}}\end{tabular} \\
AE &
  \begin{tabular}[c]{@{}c@{}}0.9967\\[-1.6mm] {\scriptsize {[}0.99, 1.00{]}}\end{tabular} &
  \begin{tabular}[c]{@{}c@{}}0.9971\\[-1.6mm] {\scriptsize {[}0.99, 1.00{]}}\end{tabular} &
  \begin{tabular}[c]{@{}c@{}}0.9909\\[-1.6mm] {\scriptsize {[}0.98, 1.00{]}}\end{tabular} &
  \begin{tabular}[c]{@{}c@{}}0.9921\\[-1.6mm] {\scriptsize {[}0.98, 1.00{]}}\end{tabular} &
  \begin{tabular}[c]{@{}c@{}}0.9575\\[-1.6mm] {\scriptsize {[}0.96, 0.96{]}}\end{tabular} &
  \textbf{\begin{tabular}[c]{@{}c@{}}0.8597\\[-1.6mm] {\scriptsize {[}0.79, 0.92{]}}\end{tabular}} &
  \begin{tabular}[c]{@{}c@{}}0.8270\\[-1.6mm] {\scriptsize {[}0.73, 0.90{]}}\end{tabular} &
  \begin{tabular}[c]{@{}c@{}}0.7934\\[-1.6mm] {\scriptsize {[}0.70, 0.88{]}}\end{tabular} &
  \begin{tabular}[c]{@{}c@{}}0.8013\\[-1.6mm] {\scriptsize {[}0.70, 0.88{]}}\end{tabular} \\
SVM-r &
  \begin{tabular}[c]{@{}c@{}}0.9967\\[-1.6mm] {\scriptsize {[}0.99, 1.00{]}}\end{tabular} &
  \begin{tabular}[c]{@{}c@{}}0.9971\\[-1.6mm] {\scriptsize {[}0.99, 1.00{]}}\end{tabular} &
  \begin{tabular}[c]{@{}c@{}}0.9794\\[-1.6mm] {\scriptsize {[}0.95, 1.00{]}}\end{tabular} &
  \begin{tabular}[c]{@{}c@{}}0.9846\\[-1.6mm] {\scriptsize {[}0.96, 1.00{]}}\end{tabular} &
  \begin{tabular}[c]{@{}c@{}}0.9464\\[-1.6mm] {\scriptsize {[}0.95, 0.95{]}}\end{tabular} &
  \begin{tabular}[c]{@{}c@{}}0.8005\\[-1.6mm] {\scriptsize {[}0.71, 0.87{]}}\end{tabular} &
  \begin{tabular}[c]{@{}c@{}}0.7473\\[-1.6mm] {\scriptsize {[}0.62, 0.86{]}}\end{tabular} &
  \begin{tabular}[c]{@{}c@{}}0.7457\\[-1.6mm] {\scriptsize {[}0.64, 0.84{]}}\end{tabular} &
  \begin{tabular}[c]{@{}c@{}}0.7739\\[-1.6mm] {\scriptsize {[}0.67, 0.86{]}}\end{tabular} \\
SVM-a &
  \begin{tabular}[c]{@{}c@{}}0.9967\\[-1.6mm] {\scriptsize {[}0.99, 1.00{]}}\end{tabular} &
  \begin{tabular}[c]{@{}c@{}}0.9971\\[-1.6mm] {\scriptsize {[}0.99, 1.00{]}}\end{tabular} &
  \begin{tabular}[c]{@{}c@{}}0.9782\\[-1.6mm] {\scriptsize {[}0.94, 1.00{]}}\end{tabular} &
  \begin{tabular}[c]{@{}c@{}}0.9837\\[-1.6mm] {\scriptsize {[}0.96, 1.00{]}}\end{tabular} &
  \begin{tabular}[c]{@{}c@{}}0.9456\\[-1.6mm] {\scriptsize {[}0.95, 0.95{]}}\end{tabular} &
  \begin{tabular}[c]{@{}c@{}}0.8135\\[-1.6mm] {\scriptsize {[}0.73, 0.88{]}}\end{tabular} &
  \begin{tabular}[c]{@{}c@{}}0.7857\\[-1.6mm] {\scriptsize {[}0.67, 0.87{]}}\end{tabular} &
  \begin{tabular}[c]{@{}c@{}}0.7454\\[-1.6mm] {\scriptsize {[}0.64, 0.84{]}}\end{tabular} &
  \begin{tabular}[c]{@{}c@{}}0.7716\\[-1.6mm] {\scriptsize {[}0.67, 0.86{]}}\end{tabular} \\
GMM-r &
  \textbf{\begin{tabular}[c]{@{}c@{}}0.9970\\[-1.6mm] {\scriptsize {[}0.99, 1.00{]}}\end{tabular}} &
  \textbf{\begin{tabular}[c]{@{}c@{}}0.9974\\[-1.6mm] {\scriptsize {[}0.99, 1.00{]}}\end{tabular}} &
  \textbf{\begin{tabular}[c]{@{}c@{}}0.9915\\[-1.6mm] {\scriptsize {[}0.98, 1.00{]}}\end{tabular}} &
  \textbf{\begin{tabular}[c]{@{}c@{}}0.9923\\[-1.6mm] {\scriptsize {[}0.98, 1.00{]}}\end{tabular}} &
  \begin{tabular}[c]{@{}c@{}}0.9566\\[-1.6mm] {\scriptsize {[}0.96, 0.96{]}}\end{tabular} &
  \begin{tabular}[c]{@{}c@{}}0.8577\\[-1.6mm] {\scriptsize {[}0.79, 0.92{]}}\end{tabular} &
  \begin{tabular}[c]{@{}c@{}}0.8071\\[-1.6mm] {\scriptsize {[}0.70, 0.90{]}}\end{tabular} &
  \begin{tabular}[c]{@{}c@{}}0.7941\\[-1.6mm] {\scriptsize {[}0.70, 0.88{]}}\end{tabular} &
  \textbf{\begin{tabular}[c]{@{}c@{}}0.8013\\[-1.6mm] {\scriptsize {[}0.71, 0.88{]}}\end{tabular}} \\
GMM-a &
  \begin{tabular}[c]{@{}c@{}}0.9964\\[-1.6mm] {\scriptsize {[}0.99, 1.00{]}}\end{tabular} &
  \begin{tabular}[c]{@{}c@{}}0.9969\\[-1.6mm] {\scriptsize {[}0.99, 1.00{]}}\end{tabular} &
  \begin{tabular}[c]{@{}c@{}}0.9909\\[-1.6mm] {\scriptsize {[}0.98, 1.00{]}}\end{tabular} &
  \begin{tabular}[c]{@{}c@{}}0.9919\\[-1.6mm] {\scriptsize {[}0.98, 1.00{]}}\end{tabular} &
  \begin{tabular}[c]{@{}c@{}}0.9563\\[-1.6mm] {\scriptsize {[}0.96, 0.96{]}}\end{tabular} &
  \begin{tabular}[c]{@{}c@{}}0.8582\\[-1.6mm] {\scriptsize {[}0.79, 0.92{]}}\end{tabular} &
  \begin{tabular}[c]{@{}c@{}}0.8073\\[-1.6mm] {\scriptsize {[}0.69, 0.90{]}}\end{tabular} &
  \textbf{\begin{tabular}[c]{@{}c@{}}0.7959\\[-1.6mm] {\scriptsize {[}0.71, 0.88{]}}\end{tabular}} &
  \begin{tabular}[c]{@{}c@{}}0.8011\\[-1.6mm] {\scriptsize {[}0.70, 0.88{]}}\end{tabular} \\ \hline
\end{tabular}
    }
\caption{AD performance with different anomaly scoring methods.}
\label{tab:tab3}
\end{minipage}
\end{table*}
\subsection{Main results}
\cref{tab:tab1} demonstrates AD performance on two datasets using Ano-NAViLa and seven competing models. First, patch-level AD performance was assessed as the models were trained, validated, and tested on \textbf{GastricLN}. Ano-NAViLa achieved a superior performance of 0.9681 AUROC, substantially outperforming its competitors.
Next, the trained models were used to evaluate WSI-level AD performance. Ano-NAViLa attained 0.9967 and 0.9894 AUROCs and 0.9971 and 0.9904 AUPRs using $A^{\max}_{score}$ and $A^{\text{top1\%}}_{score}$, respectively. For both patch-level and WSI-level AD, STFPM was the second-best model, followed by CFA and Fastflow. 
These competing models fell short by 0.0097--0.0666 in AUROC and 0.0190--0.1143 in AUPR across the two anomaly scores.
Finally, as the models were applied to \textbf{Camelyon16}, we observed a large performance gap between Ano-NAViLa and other models. Ano-NAViLa attained substantial gains of 0.1436--0.3778 in AUROC and 0.1447--0.4297 in AUPR for $A^{\max}_{score}$ and 0.0347--0.2689 in AUROC and 0.0903--0.3470 in AUPR for $A^{\text{top1\%}}_{score}$. The performance gains by Ano-NAViLa were even more pronounced on \textbf{Camelyon16}$_{\textbf{macro}}$, with AUROCs of 0.9858 and 0.9761, and AUPRs of 0.9547 and 0.9699 using $A^{\max}_{score}$ and $A^{\text{top1\%}}_{score}$, respectively, detecting macro-metastases almost perfectly and significantly outperforming other methods. The performance margins were even greater, such as 0.2148--0.4472 in AUROC and 0.3233--0.7288 in AUPR using $A^{\max}_{score}$. It is worth noting that \textbf{GastricLN} and \textbf{Camelyon16} contain different organ types (i.e., gastric and breast) acquired from different institutions, likely subject to domain shifts. Nonetheless, Ano-NAViLa demonstrated superior accuracy and robustness to domain shifts across both organ types and institutions, highlighting its potential for clinical translation and application.

\cref{tab:tab2} shows the computational efficiency of Ano-NAViLa and other competing models, measured using a batch size of 28. The metrics include the number of trainable parameters (in millions, M), latency (time required for a model to process an image and generate the anomaly score, in milliseconds, ms), and peak GPU memory usage (in mebibytes, MiB). 
Ano-NAViLa was the most efficient model with respect to the number of trainable parameters, while it ranked second in latency and fourth in memory requirement, largely due to the usage of a VLM. Hence, optimizing the VLM could further improve the computational efficiency of Ano-NAViLa.
\begin{figure*}
\centering
\includegraphics[width=1.0\textwidth,keepaspectratio]{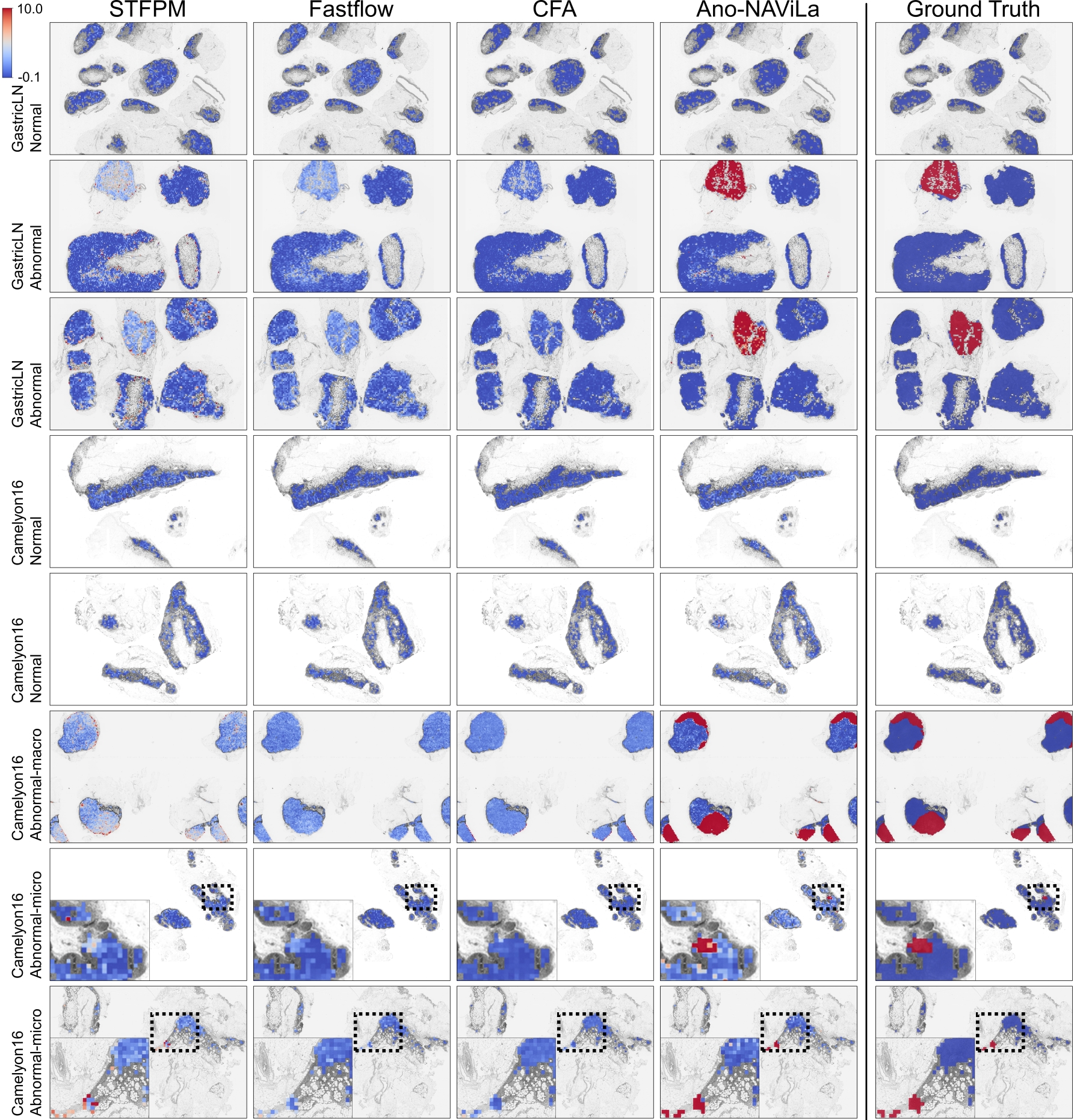}
\caption{Visualization of anomaly localization in WSIs.}
\label{fig3}
\end{figure*}

We visualize the WSI-level AD results as heatmaps in \cref{fig3} and Fig. S1 (see supplementary material). The visualization includes Ano-NAViLa, three other models with the highest AD performance, and ground truth images, with metastatic regions highlighted in red.
Due to the differences in scoring schemes and scales, we normalized the original anomaly scores using z-scores based on the mean and standard deviation of 58 normal WSIs in the test set of \textbf{GastricLN}.
Ano-NAViLa clearly distinguishes normal and abnormal regions, closely matching ground truth annotations across both datasets (\textbf{GastricLN} and \textbf{Camelyon16}). Most of metastatic regions are highlighted with higher anomaly scores, facilitating precise localization of metastatic regions in WSIs.
However, other models struggled to detect metastatic regions, producing either less confident or overly specific heatmaps. For example, in the second row of \cref{fig3}, though these models assign slightly higher scores to the metastatic regions than to normal regions, the distinction between the two regions was not as clear as in Ano-NAViLa, posing a risk of missing the metastatic regions. In the sixth row of \cref{fig3}, these models primarily highlight only the boundaries of metastatic regions, i.e., missing majority of metastatic regions, while also highlighting the boundaries of normal tissues, leading to false positives. These observations further strengthen the superior ability of Ano-NAViLa, particularly on \textbf{Camelyon16}. 

\begin{table*}[]
\centering
\resizebox{\textwidth}{!}{%
\begin{tabular}{|l|cccc|c|cccc|cccc|}
\hline
\multirow{2}{*}{Embeddings} &
  \multicolumn{4}{c|}{GastricLN (WSI)} &
  Patch &
  \multicolumn{4}{c|}{Camelyon16} &
  \multicolumn{4}{c|}{Camelyon16$_{\text{macro}}$} \\ \cline{2-14} 
 &
  \begin{tabular}[c]{@{}c@{}}AUROC\\ ($A^{\max}_{score}$)\end{tabular} &
  \begin{tabular}[c]{@{}c@{}}AUPR\\ ($A^{\max}_{score}$)\end{tabular} &
  \begin{tabular}[c]{@{}c@{}}AUROC\\ ($A^{\text{top1\%}}_{score}$)\end{tabular} &
  \begin{tabular}[c]{@{}c@{}}AUPR\\ ($A^{\text{top1\%}}_{score}$)\end{tabular} &
  AUROC &
  \begin{tabular}[c]{@{}c@{}}AUROC\\ ($A^{\max}_{score}$)\end{tabular} &
  \begin{tabular}[c]{@{}c@{}}AUPR\\ ($A^{\max}_{score}$)\end{tabular} &
  \begin{tabular}[c]{@{}c@{}}AUROC\\ ($A^{\text{top1\%}}_{score}$)\end{tabular} &
  \begin{tabular}[c]{@{}c@{}}AUPR\\ ($A^{\text{top1\%}}_{score}$)\end{tabular} &
  \begin{tabular}[c]{@{}c@{}}AUROC\\ ($A^{\max}_{score}$)\end{tabular} &
  \begin{tabular}[c]{@{}c@{}}AUPR\\ ($A^{\max}_{score}$)\end{tabular} &
  \begin{tabular}[c]{@{}c@{}}AUROC\\ ($A^{\text{top1\%}}_{score}$)\end{tabular} &
  \begin{tabular}[c]{@{}c@{}}AUPR\\ ($A^{\text{top1\%}}_{score}$)\end{tabular} \\ \hline
$\mathbf{v}^N$, $\mathbf{v}^A$ &
  \begin{tabular}[c]{@{}c@{}}0.9749\\[-1.6mm] {\scriptsize {[}0.94, 1.00{]}}\end{tabular} &
  \begin{tabular}[c]{@{}c@{}}0.9748\\[-1.6mm] {\scriptsize {[}0.94, 1.00{]}}\end{tabular} &
  \begin{tabular}[c]{@{}c@{}}0.9685\\[-1.6mm] {\scriptsize {[}0.94, 1.00{]}}\end{tabular} &
  \begin{tabular}[c]{@{}c@{}}0.9717\\[-1.6mm] {\scriptsize {[}0.94, 1.00{]}}\end{tabular} &
  \begin{tabular}[c]{@{}c@{}}0.8309\\[-1.6mm] {\scriptsize {[}0.83, 0.83{]}}\end{tabular} &
  \begin{tabular}[c]{@{}c@{}}0.7292\\[-1.6mm] {\scriptsize {[}0.63, 0.82{]}}\end{tabular} &
  \begin{tabular}[c]{@{}c@{}}0.6164\\[-1.6mm] {\scriptsize {[}0.48, 0.76{]}}\end{tabular} &
  \begin{tabular}[c]{@{}c@{}}0.7411\\[-1.6mm] {\scriptsize {[}0.65, 0.83{]}}\end{tabular} &
  \begin{tabular}[c]{@{}c@{}}0.7085\\[-1.6mm] {\scriptsize {[}0.58, 0.81{]}}\end{tabular} &
  \begin{tabular}[c]{@{}c@{}}0.7855\\[-1.6mm] {\scriptsize {[}0.66, 0.90{]}}\end{tabular} &
  \begin{tabular}[c]{@{}c@{}}0.5705\\[-1.6mm] {\scriptsize {[}0.35, 0.78{]}}\end{tabular} &
  \begin{tabular}[c]{@{}c@{}}0.8903\\[-1.6mm] {\scriptsize {[}0.79, 0.97{]}}\end{tabular} &
  \begin{tabular}[c]{@{}c@{}}0.7968\\[-1.6mm] {\scriptsize {[}0.62, 0.92{]}}\end{tabular} \\
$\mathbf{v}^I$ &
  \begin{tabular}[c]{@{}c@{}}0.9902\\[-1.6mm] {\scriptsize {[}0.97, 1.00{]}}\end{tabular} &
  \begin{tabular}[c]{@{}c@{}}0.9922\\[-1.6mm] {\scriptsize {[}0.98, 1.00{]}}\end{tabular} &
  \begin{tabular}[c]{@{}c@{}}0.9828\\[-1.6mm] {\scriptsize {[}0.96, 1.00{]}}\end{tabular} &
  \begin{tabular}[c]{@{}c@{}}0.9844\\[-1.6mm] {\scriptsize {[}0.96, 1.00{]}}\end{tabular} &
  \begin{tabular}[c]{@{}c@{}}0.9651\\[-1.6mm] {\scriptsize {[}0.96, 0.97{]}}\end{tabular} &
  \begin{tabular}[c]{@{}c@{}}0.7857\\[-1.6mm] {\scriptsize {[}0.70, 0.86{]}}\end{tabular} &
  \begin{tabular}[c]{@{}c@{}}0.6844\\[-1.6mm] {\scriptsize {[}0.54, 0.80{]}}\end{tabular} &
  \textbf{\begin{tabular}[c]{@{}c@{}}0.7898\\[-1.6mm] {\scriptsize {[}0.70, 0.87{]}}\end{tabular}} &
  \begin{tabular}[c]{@{}c@{}}0.7569\\[-1.6mm] {\scriptsize {[}0.63, 0.86{]}}\end{tabular} &
  \begin{tabular}[c]{@{}c@{}}0.8940\\[-1.6mm] {\scriptsize {[}0.83, 0.95{]}}\end{tabular} &
  \begin{tabular}[c]{@{}c@{}}0.6931\\[-1.6mm] {\scriptsize {[}0.48, 0.85{]}}\end{tabular} &
  \begin{tabular}[c]{@{}c@{}}0.9557\\[-1.6mm] {\scriptsize {[}0.89, 0.99{]}}\end{tabular} &
  \begin{tabular}[c]{@{}c@{}}0.8847\\[-1.6mm] {\scriptsize {[}0.73, 0.98{]}}\end{tabular} \\
$\mathbf{v}^I$, $\mathbf{v}^N$ &
  \begin{tabular}[c]{@{}c@{}}0.9799\\[-1.6mm] {\scriptsize {[}0.96, 1.00{]}}\end{tabular} &
  \begin{tabular}[c]{@{}c@{}}0.9806\\[-1.6mm] {\scriptsize {[}0.96, 1.00{]}}\end{tabular} &
  \begin{tabular}[c]{@{}c@{}}0.9704\\[-1.6mm] {\scriptsize {[}0.94, 1.00{]}}\end{tabular} &
  \begin{tabular}[c]{@{}c@{}}0.9698\\[-1.6mm] {\scriptsize {[}0.94, 1.00{]}}\end{tabular} &
  \begin{tabular}[c]{@{}c@{}}0.9515\\[-1.6mm] {\scriptsize {[}0.95, 0.95{]}}\end{tabular} &
  \begin{tabular}[c]{@{}c@{}}0.7980\\[-1.6mm] {\scriptsize {[}0.71, 0.88{]}}\end{tabular} &
  \begin{tabular}[c]{@{}c@{}}0.7187\\[-1.6mm] {\scriptsize {[}0.58, 0.84{]}}\end{tabular} &
  \begin{tabular}[c]{@{}c@{}}0.7667\\[-1.6mm] {\scriptsize {[}0.68 0.85{]}}\end{tabular} &
  \begin{tabular}[c]{@{}c@{}}0.7667\\[-1.6mm] {\scriptsize {[}0.66, 0.86{]}}\end{tabular} &
  \begin{tabular}[c]{@{}c@{}}0.9043\\[-1.6mm] {\scriptsize {[}0.83, 0.96{]}}\end{tabular} &
  \begin{tabular}[c]{@{}c@{}}0.7196\\[-1.6mm] {\scriptsize {[}0.51, 0.88{]}}\end{tabular} &
  \begin{tabular}[c]{@{}c@{}}0.9636\\[-1.6mm] {\scriptsize {[}0.92, 1.00{]}}\end{tabular} &
  \begin{tabular}[c]{@{}c@{}}0.9222\\[-1.6mm] {\scriptsize {[}0.82, 0.99{]}}\end{tabular} \\
$\mathbf{v}^I$, $\mathbf{v}^N$, $\mathbf{v}^A$ (Ours) &
  \textbf{\begin{tabular}[c]{@{}c@{}}0.9967\\[-1.6mm] {\scriptsize {[}0.99, 1.00{]}}\end{tabular}} &
  \textbf{\begin{tabular}[c]{@{}c@{}}0.9971\\[-1.6mm] {\scriptsize {[}0.99, 1.00{]}}\end{tabular}} &
  \textbf{\begin{tabular}[c]{@{}c@{}}0.9894\\[-1.6mm] {\scriptsize {[}0.97, 1.00{]}}\end{tabular}} &
  \textbf{\begin{tabular}[c]{@{}c@{}}0.9904\\[-1.6mm] {\scriptsize {[}0.98, 1.00{]}}\end{tabular}} &
  \textbf{\begin{tabular}[c]{@{}c@{}}0.9681\\[-1.6mm] {\scriptsize {[}0.97, 0.97{]}}\end{tabular}} &
  \textbf{\begin{tabular}[c]{@{}c@{}}0.8594\\[-1.6mm] {\scriptsize {[}0.79, 0.92{]}}\end{tabular}} &
  \textbf{\begin{tabular}[c]{@{}c@{}}0.8309\\[-1.6mm] {\scriptsize {[}0.73, 0.91{]}}\end{tabular}} &
  \begin{tabular}[c]{@{}c@{}}0.7702\\[-1.6mm] {\scriptsize {[}0.67, 0.86{]}}\end{tabular} &
  \textbf{\begin{tabular}[c]{@{}c@{}}0.7941\\[-1.6mm] {\scriptsize {[}0.69, 0.88{]}}\end{tabular}} &
  \textbf{\begin{tabular}[c]{@{}c@{}}0.9858\\[-1.6mm] {\scriptsize {[}0.96, 1.00{]}}\end{tabular}} &
  \textbf{\begin{tabular}[c]{@{}c@{}}0.9547\\[-1.6mm] {\scriptsize {[}0.88, 1.00{]}}\end{tabular}} &
  \textbf{\begin{tabular}[c]{@{}c@{}}0.9761\\[-1.6mm] {\scriptsize {[}0.92, 1.00{]}}\end{tabular}} &
  \textbf{\begin{tabular}[c]{@{}c@{}}0.9699\\[-1.6mm] {\scriptsize {[}0.89, 1.00{]}}\end{tabular}} \\ \hline
\end{tabular}
}
\caption{Results of ablation study.}
\label{tab:Ablation_O}
\end{table*}

\subsection{Ablation study}
We conduct ablation studies from two perspectives: 1) Composition of text-augmented image embeddings; 2) Generation of anomaly scores. 

\textbf{Composition of text-augmented image embeddings.} 
To analyze the effect of text-augmented image embeddings on AD performance, we conducted experiments using four different combinations of the image embedding $\mathbf{v}^I$ and two text embeddings $\mathbf{v}^N$ and $\mathbf{v}^A$: 1) text embeddings only: $\mathbf{v}^N$ and $\mathbf{v}^A$; 2) image embedding only: $\mathbf{v}^I$; 3) image embedding and normal text embedding: $\mathbf{v}^I$ and $\mathbf{v}^N$; 4) image embedding and both text embeddings: $\mathbf{v}^I$, $\mathbf{v}^N$, and $\mathbf{v}^A$ (Ours). 
\cref{tab:Ablation_O} presents the results of these four experiments. The absence of any of the three embeddings resulted in a performance drop in all cases, except for AUROC using $\mathbf{v}^I$ and $A^{\text{top1\%}}_{score}$ on \textbf{Camelyon16}. Among the three embeddings, the image embedding $\mathbf{v}^I$ had the greatest impact on AD performance, as using text embeddings only resulted in the largest performance drop. While the addition of the normal text embedding $\mathbf{v}^N$ to the image embedding $\mathbf{v}^I$ was not beneficial on \textbf{GastricLN}, the addition boosted AD performance on \textbf{Camelyon16}. Furthermore, adding the abnormal text embedding $\mathbf{v}^A$ improved AD performance for both datasets. 
These findings suggest a synergistic effect between normal and abnormal text embeddings.
Notably, using text embeddings only, we obtained comparable performance to STFPM, which is the top-performing model among other competitors.

Moreover, to ensure that the performance gain does not merely originate from the use of the VLM backbone, we evaluated the zero-shot performance of CONCH using the same term pools. As shown in Table S2 (see supplementary material), the results are significantly lower than Ano-NAViLa, demonstrating that the proposed framework contributes substantially beyond the capabilities of the VLM.

\textbf{Generation of anomaly scores.}
To compute the patch-level anomaly score $A_{score}$, we simply sum the two deviation scores $D^{N}$ and $D^{A}$. To assess its effectiveness, we compared it with five other approaches, including two simple approaches, such as maximum and L2-norm of the two deviation scores, and three methods requiring additional training, namely autoencoder (AE), one-class SVM (SVM), and Gaussian mixture model (GMM). AE is trained using the entire training set of \textbf{GastricLN}, while SVM and GMM are trained using two approaches: 1) using a random subset of the training set, resulting in SVM-r and GMM-r; 2) applying AdaBoost\cite{ref_adaboost} with 10 iterations, resulting in SVM-a and GMM-a.
As shown in \cref{tab:tab3}, AD performance was not substantially dependent on a specific scoring scheme. Although there was a favorable method for each task/dataset, the difference was minimal. These findings further emphasize the strength of the text-augmented image embeddings. 

Furthermore, we repeated the entire experiments without the erosion operation for heatmap smoothing. The experimental results, provided in the supplementary material, show that Ano-NAViLa's performance remained consistent, regardless of the erosion operation. In contrast, other competing models were generally sensitive to its use. These observations further confirm the robustness of Ano-NAViLa.

\begin{figure}
\centering
\includegraphics[width=1.0\columnwidth,keepaspectratio]{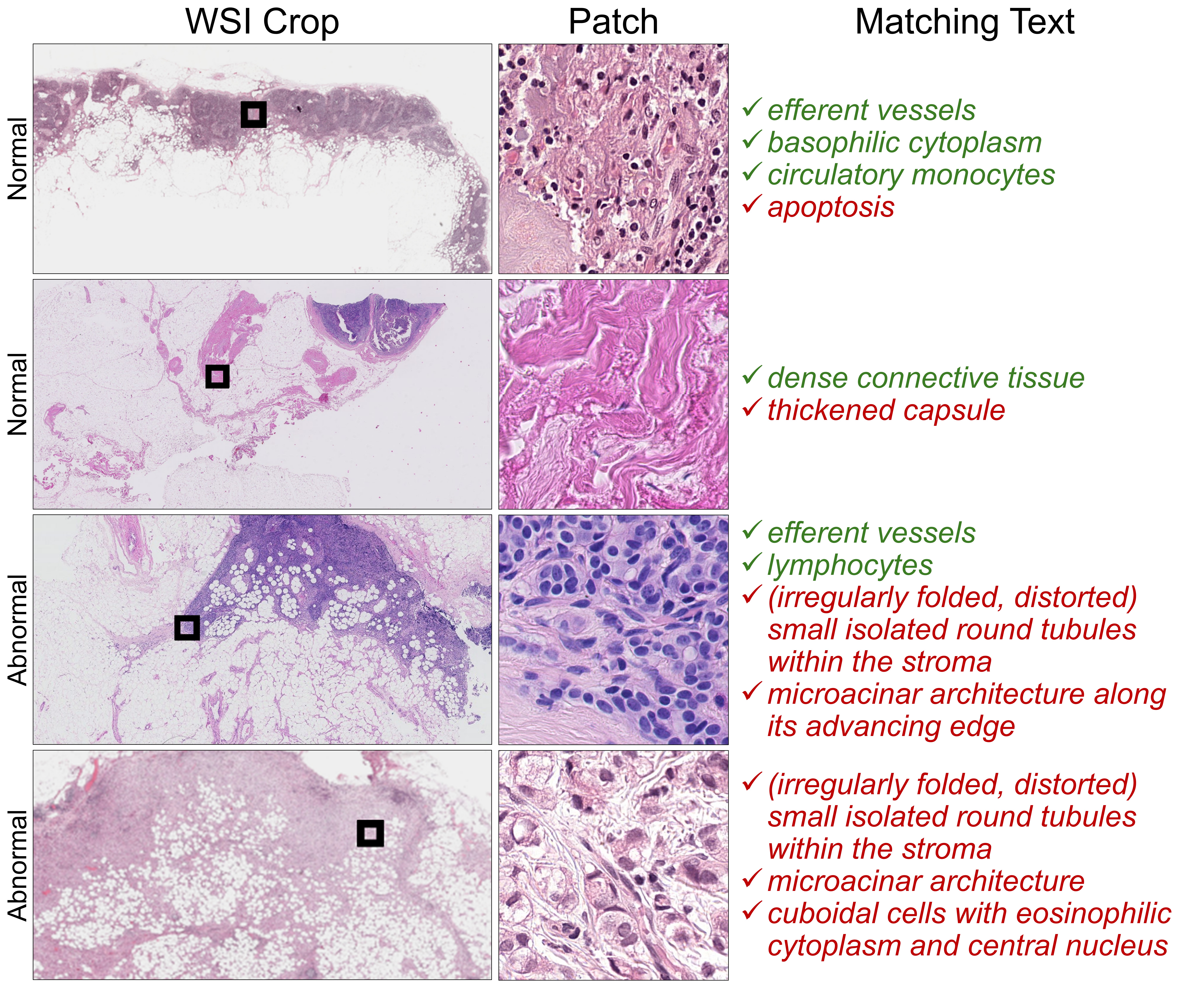}
\caption{Exemplary WSIs from Camelyon16 and their matching pathology terms. Normal and abnormal terms are shown in green and red, respectively.}
\label{fig4}
\end{figure}
\subsection{Visual and textual association and explanation}
We examined each WSI and its matching pathology terms to verify the accuracy and relevance of the image-text association through the VLM. \cref{fig4} presents exemplary \textbf{Camelyon16} WSIs, representative patches with the highest anomaly scores, and top-matching pathology terms identified by Ano-NAViLa. These selected pathology terms were reviewed and validated by an experienced pathologist. We found that the selected patches are not only representative of the WSI class label, but also that the matching pathology terms are highly relevant to both the patches and their respective class labels. These suggest that Ano-NAViLa effectively identifies relevant patches, infers disease status based on these patches, and produces histologically meaningful textual descriptions.


Moreover, we compared the distribution of image-text similarities in both \textbf{GastricLN} and \textbf{Camelyon16} as shown in Fig. S2 and S3 (see supplementary material). 
Despite differences between two datasets, the image-text similarity distributions remained consistent for patches with the same label. 
Specifically, in the normal term pool, certain terms (e.g., \textit{1. Helper T lymphocyte, 2. Small dormant lymphocytes}) exhibited higher similarity with normal-labeled patches for both datasets. Abnormal-labeled patches displayed a different distribution of similarity scores to normal patches.
Similar trends were found at the WSI-level.
Within the abnormal term pool, \textit{1. Tumor buds that emerge from medium-sized tubules} consistently showed high similarity exclusively at both patch- and WSI-levels across the two datasets. These results further support the consistent performance of Ano-NAViLa.



\section{Conclusions}
We present Ano-NAViLa, a model that detects anomalous pathology images by utilizing expert knowledge of normal and abnormal histopathology as well as data-driven knowledge from a pre-trained VLM trained on a vast amount of pathology image-text data. The experimental results demonstrate that Ano-NAViLa not only surpasses existing AD methods in performance but also enhances interpretability through pathology image-text associations, holding great potential for clinical translation and application. 
Future study will entail developing an automated protocol to construct pathology term pools for broader applications, extending validation with external datasets across various organs, and optimizing the model, particularly the VLM, to enhance computational efficiency.

\vspace{1em}
\noindent \textbf{Acknowledgments} 
This work was supported by a grant of the National Research Foundation of Korea (NRF), South Korea (No. RS-2024-00397293 and RS-2025-00558322) and Korea Institute for Advancement of Technology (KIAT) through the International Cooperative R\&D program (No. P0022543).
GastricLN dataset was collected from Korea University Anam Hospital and The Catholic University of Korea, Seoul St. Mary’s Hospital.
{
    \small
    \bibliographystyle{ieeenat_fullname}
    \bibliography{main}

\begin{thebibliography}{51}
\providecommand{\natexlab}[1]{#1}
\providecommand{\url}[1]{\texttt{#1}}
\expandafter\ifx\csname urlstyle\endcsname\relax
  \providecommand{\doi}[1]{doi: #1}\else
  \providecommand{\doi}{doi: \begingroup \urlstyle{rm}\Url}\fi

\bibitem[Akcay et~al.(2019)Akcay, Atapour-Abarghouei, and
  Breckon]{ref_ganomaly}
Samet Akcay, Amir Atapour-Abarghouei, and Toby~P. Breckon.
\newblock Ganomaly: Semi-supervised anomaly detection via adversarial training.
\newblock In \emph{Computer Vision -- ACCV 2018}, pages 622--637, 2019.

\bibitem[Batzner et~al.(2024)Batzner, Heckler, and K{\"o}nig]{ref_efficientad}
Kilian Batzner, Lars Heckler, and Rebecca K{\"o}nig.
\newblock Efficientad: Accurate visual anomaly detection at millisecond-level
  latencies.
\newblock In \emph{Proceedings of the IEEE/CVF Winter Conference on
  Applications of Computer Vision}, pages 128--138, 2024.

\bibitem[Baur et~al.(2019)Baur, Wiestler, Albarqouni, and Navab]{ref_AE1}
Christoph Baur, Benedikt Wiestler, Shadi Albarqouni, and Nassir Navab.
\newblock Deep autoencoding models for unsupervised anomaly segmentation in
  brain mr images.
\newblock In \emph{Brainlesion: Glioma, Multiple Sclerosis, Stroke and
  Traumatic Brain Injuries}, pages 161--169, 2019.

\bibitem[Bergmann et~al.(2019)Bergmann, Fauser, Sattlegger, and
  Steger]{ref_MVTecAD}
Paul Bergmann, Michael Fauser, David Sattlegger, and Carsten Steger.
\newblock Mvtec ad — a comprehensive real-world dataset for unsupervised
  anomaly detection.
\newblock In \emph{Proceedings of the IEEE/CVF Conference on Computer Vision
  and Pattern Recognition (CVPR)}, pages 9584--9592, 2019.

\bibitem[Bergmann et~al.(2020)Bergmann, Fauser, Sattlegger, and
  Steger]{ref_uninformedST}
Paul Bergmann, Michael Fauser, David Sattlegger, and Carsten Steger.
\newblock Uninformed students: Student-teacher anomaly detection with
  discriminative latent embeddings.
\newblock In \emph{Proceedings of the IEEE/CVF conference on computer vision
  and pattern recognition}, pages 4183--4192, 2020.

\bibitem[Cortes and Vapnik(1995)]{ref_SVM}
Corinna Cortes and Vladimir Vapnik.
\newblock Support-vector networks.
\newblock \emph{Machine Learning}, 20\penalty0 (3):\penalty0 273--297, 1995.

\bibitem[Defard et~al.(2021)Defard, Setkov, Loesch, and Audigier]{ref_padim}
Thomas Defard, Aleksandr Setkov, Angelique Loesch, and Romaric Audigier.
\newblock Padim: a patch distribution modeling framework for anomaly detection
  and localization.
\newblock In \emph{International Conference on Pattern Recognition}, pages
  475--489. Springer, 2021.

\bibitem[Deng and Li(2022)]{ref_reverseST}
Hanqiu Deng and Xingyu Li.
\newblock Anomaly detection via reverse distillation from one-class embedding.
\newblock In \emph{Proceedings of the IEEE/CVF conference on computer vision
  and pattern recognition}, pages 9737--9746, 2022.

\bibitem[Dippel et~al.(2024)Dippel, Preni{\ss}l, Hense, Liznerski, Winterhoff,
  Schallenberg, Kloft, Buchstab, Horst, Alber, et~al.]{ref_historare}
Jonas Dippel, Niklas Preni{\ss}l, Julius Hense, Philipp Liznerski, Tobias
  Winterhoff, Simon Schallenberg, Marius Kloft, Oliver Buchstab, David Horst,
  Maximilian Alber, et~al.
\newblock Ai-based anomaly detection for clinical-grade histopathological
  diagnostics.
\newblock \emph{NEJM AI}, 1\penalty0 (11):\penalty0 AIoa2400468, 2024.

\bibitem[Ehteshami~Bejnordi et~al.(2017)Ehteshami~Bejnordi, Veta, Johannes~van
  Diest, van Ginneken, Karssemeijer, Litjens, van~der Laak, and the
  CAMELYON16~Consortium]{ref_C16}
Babak Ehteshami~Bejnordi, Mitko Veta, Paul Johannes~van Diest, Bram van
  Ginneken, Nico Karssemeijer, Geert Litjens, Jeroen A. W.~M. van~der Laak, and
  the CAMELYON16~Consortium.
\newblock Diagnostic assessment of deep learning algorithms for detection of
  lymph node metastases in women with breast cancer.
\newblock \emph{JAMA}, 318\penalty0 (22):\penalty0 2199--2210, 2017.

\bibitem[Freund et~al.(1996)Freund, Schapire, et~al.]{ref_adaboost}
Yoav Freund, Robert~E Schapire, et~al.
\newblock Experiments with a new boosting algorithm.
\newblock In \emph{icml}, pages 148--156. Citeseer, 1996.

\bibitem[Goldstein and Hart(1999)]{ref_terms}
Neal~S. Goldstein and John Hart.
\newblock Histologic features associated with lymph node metastasis in stage t1
  and superficial t2 rectal adenocarcinomas in abdominoperineal resection
  specimens: Identifying a subset of patients for whom treatment with adjuvant
  therapy or completion abdominoperineal resection should be considered after
  local excision.
\newblock \emph{American Journal of Clinical Pathology}, 111\penalty0
  (1):\penalty0 51--58, 1999.

\bibitem[Gudovskiy et~al.(2022)Gudovskiy, Ishizaka, and Kozuka]{ref_cflow}
Denis Gudovskiy, Shun Ishizaka, and Kazuki Kozuka.
\newblock Cflow-ad: Real-time unsupervised anomaly detection with localization
  via conditional normalizing flows.
\newblock In \emph{Proceedings of the IEEE/CVF winter conference on
  applications of computer vision}, pages 98--107, 2022.

\bibitem[Huang et~al.(2023)Huang, Bianchi, Yuksekgonul, Montine, and
  Zou]{ref_PLIP}
Zhi Huang, Federico Bianchi, Mert Yuksekgonul, Thomas~J Montine, and James Zou.
\newblock A visual--language foundation model for pathology image analysis
  using medical twitter.
\newblock \emph{Nature medicine}, 29\penalty0 (9):\penalty0 2307--2316, 2023.

\bibitem[Ikezogwo et~al.(2024)Ikezogwo, Seyfioglu, Ghezloo, Geva,
  Sheikh~Mohammed, Anand, Krishna, and Shapiro]{ref_quiltnet}
Wisdom Ikezogwo, Saygin Seyfioglu, Fatemeh Ghezloo, Dylan Geva, Fatwir
  Sheikh~Mohammed, Pavan~Kumar Anand, Ranjay Krishna, and Linda Shapiro.
\newblock Quilt-1m: One million image-text pairs for histopathology.
\newblock \emph{Advances in neural information processing systems}, 36, 2024.

\bibitem[Jeong et~al.(2023)Jeong, Zou, Kim, Zhang, Ravichandran, and
  Dabeer]{ref_winclip}
Jongheon Jeong, Yang Zou, Taewan Kim, Dongqing Zhang, Avinash Ravichandran, and
  Onkar Dabeer.
\newblock Winclip: Zero-/few-shot anomaly classification and segmentation.
\newblock In \emph{Proceedings of the IEEE/CVF Conference on Computer Vision
  and Pattern Recognition}, pages 19606--19616, 2023.

\bibitem[Jia et~al.(2021)Jia, Yang, Xia, Chen, Parekh, Pham, Le, Sung, Li, and
  Duerig]{ref_align}
Chao Jia, Yinfei Yang, Ye Xia, Yi-Ting Chen, Zarana Parekh, Hieu Pham, Quoc Le,
  Yun-Hsuan Sung, Zhen Li, and Tom Duerig.
\newblock Scaling up visual and vision-language representation learning with
  noisy text supervision.
\newblock In \emph{International conference on machine learning}, pages
  4904--4916. PMLR, 2021.

\bibitem[Jiang et~al.(2022)Jiang, Tekin, Yuan, Armasu, Winham, Goode, Liu,
  Huang, Guo, and Wang]{ref_patchextraction}
Jun Jiang, Burak Tekin, Lin Yuan, Sebastian Armasu, Stacey~J Winham, Ellen~L
  Goode, Hongfang Liu, Yajue Huang, Ruifeng Guo, and Chen Wang.
\newblock Computational tumor stroma reaction evaluation led to novel
  prognosis-associated fibrosis and molecular signature discoveries in
  high-grade serous ovarian carcinoma.
\newblock \emph{Frontiers in Medicine}, 9:\penalty0 994467, 2022.

\bibitem[Komura and Ishikawa(2018)]{ref_WSIs}
Daisuke Komura and Shumpei Ishikawa.
\newblock Machine learning methods for histopathological image analysis.
\newblock \emph{Computational and structural biotechnology journal},
  16:\penalty0 34--42, 2018.

\bibitem[Lee et~al.(2022)Lee, Lee, and Song]{ref_cfa}
Sungwook Lee, Seunghyun Lee, and Byung~Cheol Song.
\newblock Cfa: Coupled-hypersphere-based feature adaptation for target-oriented
  anomaly localization.
\newblock \emph{IEEE Access}, 10:\penalty0 78446--78454, 2022.

\bibitem[Linmans et~al.(2024)Linmans, Raya, {van der Laak}, and
  Litjens]{ref_diffusionP}
Jasper Linmans, Gabriel Raya, Jeroen {van der Laak}, and Geert Litjens.
\newblock Diffusion models for out-of-distribution detection in digital
  pathology.
\newblock \emph{Medical Image Analysis}, 93:\penalty0 103088, 2024.

\bibitem[Liu et~al.(2020)Liu, Li, Zheng, Karanam, Wu, Bhanu, Radke, and
  Camps]{ref_VAE3}
Wenqian Liu, Runze Li, Meng Zheng, Srikrishna Karanam, Ziyan Wu, Bir Bhanu,
  Richard~J. Radke, and Octavia Camps.
\newblock Towards visually explaining variational autoencoders.
\newblock In \emph{2020 IEEE/CVF Conference on Computer Vision and Pattern
  Recognition (CVPR)}, pages 8639--8648, Los Alamitos, CA, USA, 2020. IEEE
  Computer Society.

\bibitem[Liznerski et~al.(2021)Liznerski, Ruff, Vandermeulen, Franks, Kloft,
  and M{\"u}ller]{ref_fcdd}
Philipp Liznerski, Lukas Ruff, Robert~A. Vandermeulen, Billy~Joe Franks, Marius
  Kloft, and Klaus-Robert M{\"u}ller.
\newblock Explainable deep one-class classification.
\newblock In \emph{International Conference on Learning Representations}, 2021.

\bibitem[Lu et~al.(2024)Lu, Chen, Williamson, Chen, Liang, Ding, Jaume,
  Odintsov, Le, Gerber, et~al.]{ref_CONCH}
Ming~Y Lu, Bowen Chen, Drew~FK Williamson, Richard~J Chen, Ivy Liang, Tong
  Ding, Guillaume Jaume, Igor Odintsov, Long~Phi Le, Georg Gerber, et~al.
\newblock A visual-language foundation model for computational pathology.
\newblock \emph{Nature Medicine}, 30:\penalty0 863--874, 2024.

\bibitem[Ngo et~al.(2019)Ngo, Winarto, Kou, Park, Akram, and Lee]{ref_GAN3}
Phuc~Cuong Ngo, Amadeus~Aristo Winarto, Connie Khor~Li Kou, Sojeong Park,
  Farhan Akram, and Hwee~Kuan Lee.
\newblock Fence gan: Towards better anomaly detection.
\newblock In \emph{2019 IEEE 31st International Conference on Tools with
  Artificial Intelligence (ICTAI)}, pages 141--148, 2019.

\bibitem[Nguyen et~al.(2024)Nguyen, Vuong, and Kwak]{ref_TQx}
Anh~Tien Nguyen, Trinh Thi~Le Vuong, and Jin~Tae Kwak.
\newblock Towards a text-based quantitative and explainable histopathology
  image analysis.
\newblock In \emph{International Conference on Medical Image Computing and
  Computer-Assisted Intervention}, pages 514--524. Springer, 2024.

\bibitem[Park et~al.(2019)Park, Liu, Wang, and Zhu]{ref_spade}
Taesung Park, Ming-Yu Liu, Ting-Chun Wang, and Jun-Yan Zhu.
\newblock Semantic image synthesis with spatially-adaptive normalization.
\newblock In \emph{Proceedings of the IEEE/CVF Conference on Computer Vision
  and Pattern Recognition (CVPR)}, 2019.

\bibitem[Radford et~al.(2021)Radford, Kim, Hallacy, Ramesh, Goh, Agarwal,
  Sastry, Askell, Mishkin, Clark, et~al.]{ref_CLIP}
Alec Radford, Jong~Wook Kim, Chris Hallacy, Aditya Ramesh, Gabriel Goh,
  Sandhini Agarwal, Girish Sastry, Amanda Askell, Pamela Mishkin, Jack Clark,
  et~al.
\newblock Learning transferable visual models from natural language
  supervision.
\newblock In \emph{International conference on machine learning}, pages
  8748--8763. PMLR, 2021.

\bibitem[Rippel et~al.(2021{\natexlab{a}})Rippel, Mertens, K{\"o}nig, and
  Merhof]{ref_GD1}
Oliver Rippel, Patrick Mertens, Eike K{\"o}nig, and Dorit Merhof.
\newblock Gaussian anomaly detection by modeling the distribution of normal
  data in pretrained deep features.
\newblock \emph{IEEE Transactions on Instrumentation and Measurement},
  70:\penalty0 1--13, 2021{\natexlab{a}}.

\bibitem[Rippel et~al.(2021{\natexlab{b}})Rippel, Mertens, and
  Merhof]{ref_MahalanobisAD}
Oliver Rippel, Patrick Mertens, and Dorit Merhof.
\newblock Modeling the distribution of normal data in pre-trained deep features
  for anomaly detection.
\newblock In \emph{2020 25th International Conference on Pattern Recognition
  (ICPR)}, pages 6726--6733. IEEE, 2021{\natexlab{b}}.

\bibitem[Roth et~al.(2022{\natexlab{a}})Roth, Pemula, Zepeda, Sch{\"o}lkopf,
  Brox, and Gehler]{ref_industry1}
Karsten Roth, Latha Pemula, Joaquin Zepeda, Bernhard Sch{\"o}lkopf, Thomas
  Brox, and Peter Gehler.
\newblock Towards total recall in industrial anomaly detection.
\newblock In \emph{Proceedings of the IEEE/CVF conference on computer vision
  and pattern recognition}, pages 14318--14328, 2022{\natexlab{a}}.

\bibitem[Roth et~al.(2022{\natexlab{b}})Roth, Pemula, Zepeda, Sch\"olkopf,
  Brox, and Gehler]{ref_patchcore}
Karsten Roth, Latha Pemula, Joaquin Zepeda, Bernhard Sch\"olkopf, Thomas Brox,
  and Peter Gehler.
\newblock Towards total recall in industrial anomaly detection.
\newblock In \emph{Proceedings of the IEEE/CVF Conference on Computer Vision
  and Pattern Recognition (CVPR)}, pages 14318--14328, 2022{\natexlab{b}}.

\bibitem[Rudolph et~al.(2022)Rudolph, Wehrbein, Rosenhahn, and
  Wandt]{ref_csflow}
Marco Rudolph, Tom Wehrbein, Bodo Rosenhahn, and Bastian Wandt.
\newblock Fully convolutional cross-scale-flows for image-based defect
  detection.
\newblock In \emph{Proceedings of the IEEE/CVF Winter Conference on
  Applications of Computer Vision}, pages 1088--1097, 2022.

\bibitem[Rudolph et~al.(2023)Rudolph, Wehrbein, Rosenhahn, and
  Wandt]{ref_asymmetricST}
Marco Rudolph, Tom Wehrbein, Bodo Rosenhahn, and Bastian Wandt.
\newblock Asymmetric student-teacher networks for industrial anomaly detection.
\newblock In \emph{Proceedings of the IEEE/CVF winter conference on
  applications of computer vision}, pages 2592--2602, 2023.

\bibitem[Salehi et~al.(2021)Salehi, Sadjadi, Baselizadeh, Rohban, and
  Rabiee]{ref_multiresolutionST}
Mohammadreza Salehi, Niousha Sadjadi, Soroosh Baselizadeh, Mohammad~H Rohban,
  and Hamid~R Rabiee.
\newblock Multiresolution knowledge distillation for anomaly detection.
\newblock In \emph{Proceedings of the IEEE/CVF conference on computer vision
  and pattern recognition}, pages 14902--14912, 2021.

\bibitem[Schlegl et~al.(2017)Schlegl, Seeböck, Waldstein, Schmidt-Erfurth, and
  Langs]{ref_GAN1}
Thomas Schlegl, Philipp Seeböck, Sebastian~M. Waldstein, Ursula
  Schmidt-Erfurth, and Georg Langs.
\newblock Unsupervised anomaly detection with generative adversarial networks
  to guide marker discovery.
\newblock In \emph{Information Processing in Medical Imaging}, pages 146--157,
  2017.

\bibitem[Sun et~al.(2018)Sun, Wang, Xiong, and Shao]{ref_VAE2}
Jiayu Sun, Xinzhou Wang, Naixue Xiong, and Jie Shao.
\newblock Learning sparse representation with variational auto-encoder for
  anomaly detection.
\newblock \emph{IEEE Access}, 6:\penalty0 33353--33361, 2018.

\bibitem[Wang et~al.(2021)Wang, Han, Ding, and Huang]{ref_stfpm}
Guodong Wang, Shumin Han, Errui Ding, and Di Huang.
\newblock Student-teacher feature pyramid matching for anomaly detection.
\newblock \emph{The British Machine Vision Conference (BMVC)}, 2021.

\bibitem[Wang et~al.(2022)Wang, Yu, Yu, Dai, Tsvetkov, and Cao]{ref_simvlm}
Zirui Wang, Jiahui Yu, Adams~Wei Yu, Zihang Dai, Yulia Tsvetkov, and Yuan Cao.
\newblock Simvlm: Simple visual language model pretraining with weak
  supervision.
\newblock In \emph{International Conference on Learning Representations}, 2022.

\bibitem[Wold et~al.(1987)Wold, Esbensen, and Geladi]{ref_PCA}
Svante Wold, Kim Esbensen, and Paul Geladi.
\newblock Principal component analysis.
\newblock \emph{Chemometrics and Intelligent Laboratory Systems}, 2\penalty0
  (1):\penalty0 37--52, 1987.
\newblock Proceedings of the Multivariate Statistical Workshop for Geologists
  and Geochemists.

\bibitem[Wu et~al.(2021)Wu, Chen, Fuh, and Liu]{ref_AE3}
Jhih-Ciang Wu, Ding-Jie Chen, Chiou-Shann Fuh, and Tyng-Luh Liu.
\newblock Learning unsupervised metaformer for anomaly detection.
\newblock In \emph{Proceedings of the IEEE/CVF International Conference on
  Computer Vision (ICCV)}, pages 4369--4378, 2021.

\bibitem[Wyatt et~al.(2022)Wyatt, Leach, Schmon, and Willcocks]{ref_anoddpm}
Julian Wyatt, Adam Leach, Sebastian~M. Schmon, and Chris~G. Willcocks.
\newblock Anoddpm: Anomaly detection with denoising diffusion probabilistic
  models using simplex noise.
\newblock In \emph{Proceedings of the IEEE/CVF Conference on Computer Vision
  and Pattern Recognition (CVPR) Workshops}, pages 650--656, 2022.

\bibitem[Yan et~al.(2023)Yan, Zhang, Liu, Pang, and Wang]{ref_diff1}
Cheng Yan, Shiyu Zhang, Yang Liu, Guansong Pang, and Wenjun Wang.
\newblock Feature prediction diffusion model for video anomaly detection.
\newblock In \emph{Proceedings of the IEEE/CVF International Conference on
  Computer Vision (ICCV)}, pages 5527--5537, 2023.

\bibitem[Yi and Yoon(2020)]{ref_patchsvdd}
Jihun Yi and Sungroh Yoon.
\newblock Patch svdd: Patch-level svdd for anomaly detection and segmentation.
\newblock In \emph{Proceedings of the Asian conference on computer vision},
  2020.

\bibitem[Yu et~al.(2021)Yu, Zheng, Wang, Li, Wu, Zhao, and Wu]{ref_fastflow}
Jiawei Yu, Ye Zheng, Xiang Wang, Wei Li, Yushuang Wu, Rui Zhao, and Liwei Wu.
\newblock Fastflow: Unsupervised anomaly detection and localization via 2d
  normalizing flows.
\newblock \emph{arXiv preprint arXiv:2111.07677}, 2021.

\bibitem[Yu et~al.(2022)Yu, Wang, Vasudevan, Yeung, Seyedhosseini, and
  Wu]{ref_coca}
Jiahui Yu, Zirui Wang, Vijay Vasudevan, Legg Yeung, Mojtaba Seyedhosseini, and
  Yonghui Wu.
\newblock Coca: Contrastive captioners are image-text foundation models.
\newblock \emph{Transactions on Machine Learning Research}, 2022.

\bibitem[Zhang et~al.(2025)Zhang, Xu, Usuyama, Xu, Bagga, Tinn, Preston, Rao,
  Wei, Valluri, Wong, Tupini, Wang, Mazzola, Shukla, Liden, Gao, Crabtree,
  Piening, Bifulco, Lungren, Naumann, Wang, and Poon]{ref_biomedclip}
Sheng Zhang, Yanbo Xu, Naoto Usuyama, Hanwen Xu, Jaspreet Bagga, Robert Tinn,
  Sam Preston, Rajesh Rao, Mu Wei, Naveen Valluri, Cliff Wong, Andrea Tupini,
  Yu Wang, Matt Mazzola, Swadheen Shukla, Lars Liden, Jianfeng Gao, Angela
  Crabtree, Brian Piening, Carlo Bifulco, Matthew~P. Lungren, Tristan Naumann,
  Sheng Wang, and Hoifung Poon.
\newblock A multimodal biomedical foundation model trained from fifteen million
  image–text pairs.
\newblock \emph{NEJM AI}, 2\penalty0 (1):\penalty0 AIoa2400640, 2025.

\bibitem[Zhou et~al.(2023)Zhou, Pang, Tian, He, and Chen]{ref_anomalyclip}
Qihang Zhou, Guansong Pang, Yu Tian, Shibo He, and Jiming Chen.
\newblock Anomalyclip: Object-agnostic prompt learning for zero-shot anomaly
  detection.
\newblock In \emph{The Twelfth International Conference on Learning
  Representations}, 2023.

\bibitem[Zhu and Pang(2024)]{ref_inctrl}
Jiawen Zhu and Guansong Pang.
\newblock Toward generalist anomaly detection via in-context residual learning
  with few-shot sample prompts.
\newblock In \emph{Proceedings of the IEEE/CVF conference on computer vision
  and pattern recognition}, pages 17826--17836, 2024.

\bibitem[Zingman et~al.(2024)Zingman, Stierstorfer, Lempp, and
  Heinemann]{ref_diverse}
Igor Zingman, Birgit Stierstorfer, Charlotte Lempp, and Fabian Heinemann.
\newblock Learning image representations for anomaly detection: application to
  discovery of histological alterations in drug development.
\newblock \emph{Medical Image Analysis}, 92:\penalty0 103067, 2024.

\bibitem[Zou et~al.(2023)Zou, Yang, Kui, Liu, Liao, and Zhao]{ref_GD2}
Beiji Zou, Kangkang Yang, Xiaoyan Kui, Jun Liu, Shenghui Liao, and Wei Zhao.
\newblock Anomaly detection for streaming data based on grid-clustering and
  gaussian distribution.
\newblock \emph{Information Sciences}, 638:\penalty0 118989, 2023.

\end{thebibliography}
}

\end{document}